\documentclass{article} % For LaTeX2e
\usepackage[preprint]{colm2026_conference}

\usepackage{microtype}
\usepackage{hyperref}
\usepackage{url}
\usepackage{booktabs}

\usepackage[ruled,vlined]{algorithm2e}

\usepackage{graphicx}
\usepackage{subcaption}
\usepackage{multirow}
\usepackage{colortbl}
\usepackage[dvipsnames, table]{xcolor}
\usepackage[most]{tcolorbox}
\usepackage{makecell}

\usepackage{amsmath}
\usepackage{amssymb}
\usepackage{mathtools}
\usepackage{amsthm}
\usepackage{ulem}
\usepackage{tabularx}
\usepackage{tikz}
\usepackage{xstring}
\usepackage{wrapfig}

\usepackage{lineno}

\definecolor{darkblue}{rgb}{0, 0, 0.5}
\hypersetup{colorlinks=true, citecolor=darkblue, linkcolor=darkblue, urlcolor=darkblue}

\title{Co-Evolution of Policy and Internal Reward for \\ Language Agents} 

\author{\mdseries
Xinyu Wang$^{1}$\thanks{Equal contribution.}, 
Hanwei Wu$^{2}$\footnotemark[1], 
Jingwei Song$^{3}$\footnotemark[1], 
Shuyuan Zhang$^{1,9}$, \\
Jiayi Zhang$^{4}$,
Fanqi Kong$^{5}$,
Tung Sum Thomas Kwok$^{6}$,
Xiao-Wen Chang$^{1}$, \\
Yuyu Luo$^{4}$, 
Chenglin Wu$^{7}$, 
Bang Liu$^{8,9}$\thanks{Corresponding author.}\\
$^{1}$McGill University, 
$^{2}$McMaster University, 
$^{3}$The University of Hong Kong, \\
$^{4}$The Hong Kong University of Science and Technology (Guangzhou), \\
$^{5}$Peking University, 
$^{6}$University of California, Los Angeles, \\
$^{7}$DeepWisdom, 
$^{8}$Université de Montréal,
$^{9}$Mila \\
\texttt{\href{mailto:xinyu.wang5@mail.mcgill.ca}{xinyu.wang5@mail.mcgill.ca}, \href{mailto:bang.liu@umontreal.ca}{bang.liu@umontreal.ca}}
}

\begin{document}

\ifcolmsubmission
\linenumbers
\fi

\maketitle

\begin{abstract}

Large language model (LLM) agents learn by interacting with environments, but long-horizon training remains fundamentally bottlenecked by sparse and delayed rewards.
Existing methods typically address this challenge through post-hoc credit assignment or external reward models, which provide limited guidance at inference time and often separate reward improvement from policy improvement.
We propose Self-Guide, a self-generated internal reward for language agents that supports both inference-time guidance and training-time supervision.
Specifically, the agent uses Self-Guide as a short self-guidance signal to steer the next action during inference, and converts the same signal into step-level internal reward for denser policy optimization during training.
This creates a co-evolving loop: better policy produces better guidance, and better guidance further improves policy as internal reward.
Across three agent benchmarks, inference-time self-guidance already yields clear gains, while jointly evolving policy and internal reward with GRPO brings further improvements (8\%) over baselines trained solely with environment reward.
Overall, our results suggest that language agents can improve not only by collecting more experience, but also by learning to generate and refine their own internal reward during acting and learning.

\end{abstract}

% 核心故事线
% Sparse outcome rewards are insufficient for long-horizon LLM agents due to poor credit assignment.
% We propose using the same LLM as an internal rater to provide semantically meaningful intermediate feedback, forming a dense internal reward. (rater provides internal reward, internal reward forms dense reward)
% The key idea is that the LLM already encodes rich world knowledge, which can be reused as a self-evaluator to generate dense reward signals without external supervision.
\newcommand{\ourmethod}{Self-Guidance}
\newcommand{\ourmethodshort}{SG}
\newcommand{\ourmethodfull}{Self-Guidance \& Guidance reward}
\newcommand{\ourmethodfullshort}{SG \& GR}

\section{Introduction}
\label{sec:intro}

% Language agents acting in long-horizon interactive environments require better guidance at both inference time and training time. In web navigation, scientific experimentation, and embodied interaction, the environment often reveals success only at the end of a long trajectory, making it unclear which intermediate decisions actually advance the task and which are redundant or harmful \citep{yao2023webshopscalablerealworldweb,wang2022scienceworldagentsmarter5th,shridhar2021alfworldaligningtextembodied}.

Humans often form internal self-guidance while acting, and later improve not only from outcomes but also from these self-generated judgments. In contrast, language agents acting in long-horizon interactive environments, such as web navigation, scientific experimentation, and embodied interaction, typically receive task feedback only at the end of a long trajectory \citep{yao2023webshopscalablerealworldweb,wang2022scienceworldagentsmarter5th,shridhar2021alfworldaligningtextembodied}. This makes it unclear which intermediate decisions actually advanced the task and which were redundant or harmful, creating a fundamental challenge for both inference-time decision making and training-time policy improvement.

% 第三段：

% 第四段：

Recent work has begun to mitigate this sparse-reward bottleneck. One line improves agent learning through post-hoc credit assignment, for example by retrospectively redistributing credit or refining rollout optimization after trajectories have been collected \citep{tan2026hindsightcreditassignmentlonghorizon, dong2026agentic,wang2025ragen, kong2026infopo}. Another line introduces explicit process-level supervision, including process reward models that estimate step-wise promise and progress for intermediate agent behavior \citep{xi2025agentprmprocessrewardmodels}. In most existing approaches, intermediate signals are used mainly after rollout for training, rather than during rollout to guide the next decision. When these signals are produced by separate evaluators or reward models, they also add overhead and can drift away from the policy's evolving rollout distribution.

\begin{figure}[t]
  \centering
  \includegraphics[width=\linewidth]{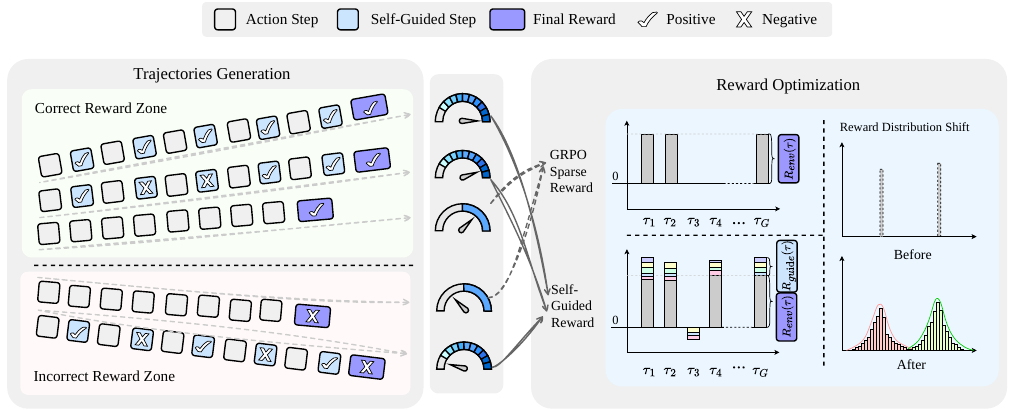}
\caption{
Sparse environment reward alone may fail to distinguish trajectories with different intermediate quality, leaving GRPO with weakly separated training signals.
Our method interleaves self-guided steps with action steps during trajectory generation, so that the same signal can guide action selection at inference time and be aggregated as internal reward at training time.
This yields denser and more discriminative trajectory-level supervision for policy optimization.
}
  \label{fig:motivation}
\end{figure}

Motivated by this gap, we propose \textbf{Self-Guide}, a self-generated internal reward for language agents. At inference time, language agent produces a short self-guidance signal on the current trajectory prefix and uses it to guide the next action. At training time, the same signal is converted into step-level internal reward, yielding denser supervision beyond sparse environment outcomes. In this way, Self-Guide serves as a single reward form that supports both acting and learning within one unified training loop.

A central challenge in applying Self-Guide to policy optimization is that self-generated reward is useful but imperfect. Unlike fixed reward shaping, our setting is one of internal co-evolution: as policy improves, it produces better guidance and the improved guidance in turn provides a stronger internal reward for further policy improvement. This makes training both promising and nontrivial: if Self-Guide is introduced too aggressively, it can amplify immature judgments early in training; and because it is not potential-based, naive reward shaping may also shift optimization away from the true environment objective. We therefore use a \textbf{stage-wise trust schedule} (Figure~\ref{fig:motivation}) that first uses Self-Guide only for inference-time guidance, then gradually activates it as internal reward, and finally anneals its contribution later in training. In this way, policy and internal reward can co-evolve under the agent's own online interaction distribution without destabilizing optimization.

Experiments on ALFWorld, ScienceWorld, and WebShop~\citep{shridhar2021alfworldaligningtextembodied,wang2022scienceworldagentsmarter5th,yao2023webshopscalablerealworldweb} show that Self-Guide improves language-agent learning at both inference time and training time. In particular, at training time, jointly optimizing policy and internal reward consistently outperforms GRPO across benchmarks. For example, with Qwen3-4B, our method achieves around \textbf{8\%} average improvement over GRPO, showing that co-evolving policy and internal reward provides a more effective training signal than relying on sparse environment reward alone. Ablation studies clarify that the gain does not come from simply adding self-generated reward. Instead, Self-Guide works best when it is first used to guide actions, then gradually promoted into internal reward as training proceeds; moreover, offline-distilled self-guidance does not transfer reliably to downstream RL, indicating that policy and internal reward must improve together online under the agent’s own rollout distribution.

Our main contributions are as follows:
\begin{itemize}
    \item We propose \textbf{Self-Guide}, a new self-generated internal reward form for language agents that can be used both as inference-time self-guidance and as training-time internal reward.
    \item We develop a \textbf{stage-wise trust schedule} that makes self-generated reward shaping effective and stable for agent RL training.
    \item We show on ALFWorld, ScienceWorld, and WebShop that Self-Guide yields significant gains at both inference time and training time, and that co-evolving policy and internal reward further improves over reward-from-environment-only baselines.
\end{itemize}

\section{Related Work}

\subsection{Agentic Reinforcement Learning}

Reinforcement learning is increasingly used to train LLM-based agents that interact with environments over multiple turns. Early efforts study multi-turn credit assignment through hierarchical training \citep{zhou2024archer, zhou2025sweet}, while subsequent work induces grounded tool use via RL-driven search and clarification \citep{jin2025search, qian2025toolrl, acikgoz2025speakrl, chen2025rl4lm}; analyses of long-rollout dynamics reveal persistent instability and motivate algorithmic stabilization \citep{wang2025arlstability}. On the optimization side, RLVR-style algorithms have been adapted for long, high-variance agent trajectories through group-relative policy gradients \citep{shao2024deepseekmath, feng2025group}, sequence-level normalization \citep{zheng2025group, yu2025dapo}, efficient sampling strategies \citep{sheng2025espo}, and relaxed synchronization constraints \citep{xu2025single}. RAGEN \citep{wang2025ragen} proposes StarPO for trajectory-level optimization in stochastic environments, identifying training collapse patterns and introducing stabilized variants. For tool-augmented reasoning, ARPO/AEPO address post-tool entropy spikes with adaptive rollout branching and advantage attribution \citep{dong2026agentic, dong2025agentic1}; for multi-reward settings, GDPO decouples reward normalization \citep{liu2026gdpo}, ARIA aggregates rewards by intention \citep{yang2025aria} and InfoPO assigns credit 
via turn-level information advantage without task-specific reward shaping or trained 
process reward models \citep{kong2026infopo}. A separate line explores co-evolutionary training: RLAnything \citep{wang2026rlanything} jointly optimizes environment, policy, and reward model in a closed loop, while GenEnv \citep{guo2025genenv} co-evolves agent and environment simulator via difficulty-aligned curriculum generation. 

\subsection{Credit Assignment in Agent Reinforcement Learning}

A central challenge in long-horizon agent RL is credit assignment under sparse, delayed rewards \citep{tan2026hindsightcreditassignmentlonghorizon, peng2026hiperhierarchicalreinforcementlearning}. 
\citet{feng2025groupingrouppolicyoptimizationllm} introduce hierarchical relative advantages for finer-grained credit, \citet{li2025saltstepleveladvantageassignment} refine step-level advantages over trajectory graphs, and \citet{chan2024dense, cao2025scar} exploit reward models for dense signals. \citet{tan2026hindsightcreditassignmentlonghorizon} propose HCAPO, leveraging the LLM as a post-hoc critic to refine step-level Q-values through hindsight reasoning, showing that value-free methods like GRPO produce inaccurate step-level estimates under reward sparsity. On the process supervision 
side, \citet{xi2025agentprmprocessrewardmodels} develops AgentPRM with TD-based estimation to evaluate agent actions by their promise and progress toward the goal.\citet{choudhury2025processrewardmodelsllm} propose a lightweight actor-critic framework using Monte Carlo rollouts to construct process reward 
targets. Beyond explicit supervision, \citet{cui2025processreinforcementimplicitrewards} induce dense process rewards from rollouts and outcome labels without separate annotation, and \citet{liu2025agenticreinforcementlearningimplicit} learn implicit step rewards from trajectory preferences. These works support our premise that outcome-only reward is too coarse for long-horizon training, though they mainly improve how externally provided rewards are redistributed across a trajectory.

\section{Method}
\label{sec:method}

Long-horizon interactive tasks provide only sparse terminal rewards, making credit assignment for intermediate decisions difficult.
We address this with two coupled uses of \textbf{Self-Guide}.
First, at each rollout step, the agent generates a verbal self-guidance signal that steers action selection at inference time (Section~\ref{sec:itsg}).
Second, the same signal is reused as dense internal reward during training (Section~\ref{sec:reward}).
Since self-guidance must mature before it can be trusted as reward, we further introduce a stage-wise trust schedule that progressively promotes guidance into internal reward as training proceeds (Section~\ref{sec:schedule}).

\subsection{Problem Setup}

Under Self-Guide, an episode is represented as
\begin{equation}
    \tau = \{(o_t, z_t, a_t)\}_{t=1}^{T},
\end{equation}
where at step $t$, the agent observes $o_t$ from the environment, produces a verbal self-guidance signal $z_t$, and then takes action $a_t$.
The history
\begin{equation}
    h_{t-1} = \{(o_i, z_i, a_i)\}_{i=1}^{t-1}
\end{equation}
captures the full interaction context up to step $t-1$.
The environment returns a sparse task-level reward $R_{\mathrm{env}}(\tau)$, typically $1$ for task success and $0$ otherwise, which provides no intermediate credit assignment across the $T$ steps of the trajectory.
Our goal is to augment this sparse setting with a self-generated signal that is useful both at inference time and as a training signal.

\subsection{Inference-Time Self-Guidance}
\label{sec:itsg}

Without intermediate feedback, the agent has no explicit basis for assessing whether its current trajectory is heading in the right direction.
We address this by having the model generate a short verbal self-guidance signal $z_t$ before each action:
\begin{equation}
z_t \sim \pi_{\theta}(\cdot \mid h_{t-1}, o_t).
\end{equation}
Here, $z_t$ is a natural-language assessment of whether the trajectory so far is making progress, has gone off-track, or is approaching a subgoal.
The model then acts conditioned on both the trajectory context and its own self-guidance:
\begin{equation}
a_t \sim \pi_{\theta}(\cdot \mid h_{t-1}, o_t, z_t).
\end{equation}

This two-stage generation forces the model to verbalize its current assessment before acting, providing explicit in-context steering before each action.
Importantly, because $z_t$ and $a_t$ are produced by the same model $\pi_\theta$, the model is continuously trained to generate guidance that is actionable for its own decision-making, progressively improving guidance quality across rollouts.
This progressive improvement is what later makes $z_t$ a reliable source of internal reward.

\begin{figure*}[t]
  \centering
  \includegraphics[width=\textwidth]{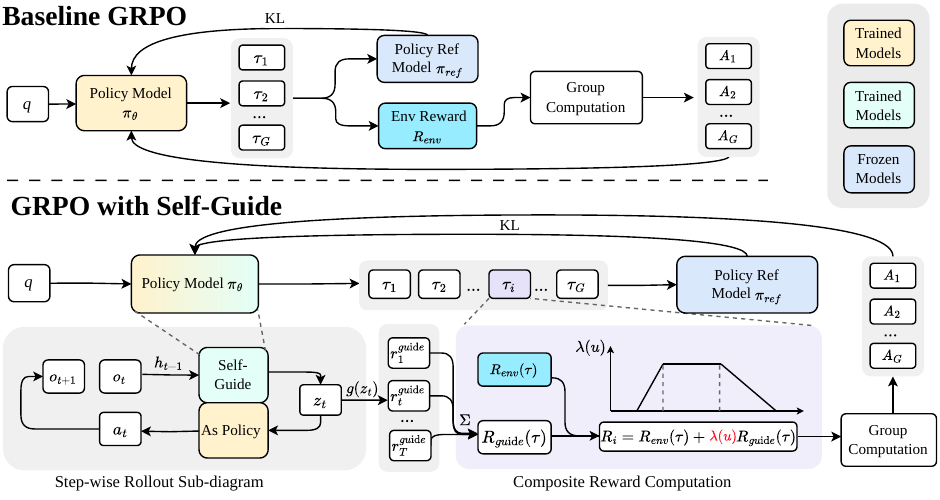}
  \caption{
  \textbf{Comparison between baseline GRPO and GRPO with Self-Guide.}
  Baseline GRPO optimizes a policy using sparse trajectory-level environment rewards.
  Our method augments each step with a verbal self-guidance signal $z_t$: the model first generates $z_t$ to assess the current trajectory, and then produces action $a_t$ conditioned on $z_t$.
  The same self-guidance signals are mapped to step-level internal rewards, aggregated into $R_{\mathrm{sg}}(\tau)$, and combined with $R_{\mathrm{env}}(\tau)$ via a stage-dependent coefficient $\lambda(u)$ for joint policy optimization.
  }
  \label{fig:main_method}
\end{figure*}

\subsection{Self-Guidance as Internal Reward}
\label{sec:reward}

Inference-time self-guidance improves acting, but $R_{\mathrm{env}}$ remains sparse.
Since $z_t$ is already generated at every step, a natural next question is whether the same signal can also provide dense, step-level credit assignment during training.

We realize this second role by converting the verbal self-guidance signal into a scalar reward:
\begin{equation}
r_t^{\mathrm{sg}} = g(z_t),
\end{equation}
where $g(\cdot)$ maps the verbal assessment to a scalar reflecting the local contribution of the current step toward task completion.
In practice, we implement $g$ as a discrete polarity mapping:
\emph{positive}, \emph{neutral}, and \emph{negative} assessments are mapped to $+0.1$, $0$, and $-0.1$, respectively.
We find this sufficient to capture meaningful step-level progress signals without introducing additional scoring complexity.

The environment reward and the aggregated self-guidance reward are linearly combined into a unified trajectory-level training signal:
\begin{equation}
R(\tau; u) = R_{\mathrm{env}}(\tau) + \lambda(u)\sum_{t=1}^{T} r_t^{\mathrm{sg}},
\end{equation}
where $u$ denotes the current training stage and $\lambda(u)\in[0,1]$ controls the contribution of the internal reward.
This dual-purpose design, where the same $z_t$ steers inference-time action selection and shapes training-time reward, creates a self-reinforcing loop:
a stronger policy generates more coherent trajectories, which produce more reliable self-guidance signals, which in turn yield more accurate internal rewards and further strengthen the policy.

\subsection{Joint Optimization with Self-Guide}

With the composite reward $R(\tau;u)$ defined, we jointly optimize the policy and Self-Guide with a single GRPO objective.
For each environment instance, we sample a group of rollouts $\{\tau_i\}_{i=1}^{G}$ from the current model, each receiving the combined return
\begin{equation}
R_i = R(\tau_i;u).
\end{equation}
Group-relative advantages are then computed as
\begin{equation}
\hat{A}_i = \frac{R_i - \mu_R}{\sigma_R + \epsilon},
\end{equation}
where $\mu_R$ and $\sigma_R$ are the mean and standard deviation of returns within the group.
The GRPO objective is
\begin{equation}
\mathcal{L}_{\pi}(\theta)
=
\mathbb{E}\left[
\min\left(
r_i(\theta)\hat{A}_i,\,
\mathrm{clip}(r_i(\theta), 1-\varepsilon, 1+\varepsilon)\hat{A}_i
\right)
\right],
\end{equation}
where $r_i(\theta)$ is the probability ratio between the updated model and the reference model, and $\varepsilon$ is the clipping coefficient.

Since both $z_t$ and $a_t$ are generated by the same $\pi_\theta$, this single objective simultaneously trains the model to produce better self-guidance and to act better conditioned on that guidance.
In other words, the co-evolution of policy and internal reward is realized end-to-end through one unified training signal.
The remaining question is how $\lambda(u)$ should be scheduled across training, which we address next.

The composite reward $R(\tau;u)$ is not specific to GRPO and is compatible with other group-relative or advantage-based objectives.
We verify this with DAPO, confirming that the gains from Self-Guide transfer across optimizers (Section~\ref{sec:ablation_optimizer}).

\subsection{Stage-Wise Trust Schedule}
\label{sec:schedule}

The co-evolving loop introduced above faces a fundamental bootstrap problem:
reliable internal reward requires mature self-guidance, yet mature self-guidance emerges only through sufficient training.
Naively applying internal reward from the start forces the policy to optimize against its own immature signals, which can destabilize learning.
We resolve this chicken-and-egg tension with a trapezoid trust schedule $\lambda(u)$ that governs when and how strongly self-guidance contributes as internal reward:
\begin{equation}
\lambda(u) = 
\begin{cases}
0 & \text{Phase I: guidance-only warm-up,} \\
0 \rightarrow 1 & \text{Phase II: reward activation,} \\
1 & \text{Phase III: full internal reward,} \\
1 \rightarrow 0 & \text{Phase IV: late annealing.}
\end{cases}
\end{equation}

The key design principle is that trust follows competence:
internal reward is activated only after self-guidance has demonstrated sufficient reliability, and is later attenuated to ensure that the final policy remains faithful to the true environment objective.

\paragraph{Phase I: Guidance-only warm-up.}
Self-guidance conditions each action but contributes no reward ($\lambda=0$).
This decouples the two roles of $z_t$: the model learns to generate and act on guidance purely through environmental feedback, without being simultaneously asked to treat its own immature assessments as a training signal.
By the end of this phase, guidance quality has stabilized sufficiently to serve as a meaningful source of internal reward.

\paragraph{Phase II: Reward activation.}
$\lambda(u)$ is smoothly ramped from $0$ to $1$, gradually promoting self-guidance into internal reward.
The linear ramp avoids the optimization shock of abruptly introducing a new reward component, giving the policy time to recalibrate around the denser signal.

\paragraph{Phase III: Full internal reward.}
With both roles of $z_t$ mutually calibrated, the co-evolving loop operates at full strength:
dense step-level credit from self-guidance complements sparse environment outcomes, and each improvement in one reinforces the other.

\paragraph{Phase IV: Late annealing.}
$\lambda$ is gradually reduced from $1$ to $0$, progressively withdrawing the internal reward.
Because the self-guidance-derived reward is not potential-based~\citep{ng1999policy}, retaining it indefinitely would bias the converged policy away from the true environment objective.
Late annealing ensures that the terminal policy is ultimately optimized under $R_{\mathrm{env}}$, preserving alignment with task-level goals.

\medskip
\noindent
The specific transition points are soft hyperparameters:
the essential structure---warm-up before activation, and annealing before convergence---matters more than the precise timing.
We verify this robustness in Section~\ref{sec:ablation_schedule_timing}.
\definecolor{OurLightGray}{RGB}{242,244,247}
\definecolor{OurDarkGray}{RGB}{224,228,235}
\newcommand{\hll}{\cellcolor{OurLightGray}}
\newcommand{\hld}{\cellcolor{OurDarkGray}}

\definecolor{DeltaBlue}{RGB}{47,111,237}
\definecolor{DeltaBlueDark}{RGB}{34,84,180}
\definecolor{DeltaBlueNeg}{RGB}{92,118,168}

% max bar width and height
\newcommand{\maxbarwidth}{0.55}
\newcommand{\barheight}{0.11}

% normalization constant: adjust based on your typical delta range
\newcommand{\deltamax}{30}

% scaled positive bar
\newcommand{\incLite}[1]{%
{\scriptsize\textcolor{DeltaBlue}{\,+#1}}%
\,\tikz[baseline=-0.55ex]{
  \pgfmathsetmacro{\w}{min(\maxbarwidth, \maxbarwidth*(#1)/\deltamax)}
  \fill[DeltaBlue!65] (0,0) rectangle (\w cm,\barheight cm);
}}

\newcommand{\incDeep}[1]{%
{\scriptsize\textcolor{DeltaBlueDark}{\,+#1}}%
\,\tikz[baseline=-0.55ex]{
  \pgfmathsetmacro{\w}{min(\maxbarwidth, \maxbarwidth*(#1)/\deltamax)}
  \fill[DeltaBlueDark] (0,0) rectangle (\w cm,\barheight cm);
}}

% scaled negative bar: pass absolute value, display minus manually
\newcommand{\decLite}[1]{%
{\scriptsize\textcolor{DeltaBlueNeg}{\,#1}}%
\,\tikz[baseline=-0.55ex]{
  \pgfmathsetmacro{\absval}{abs(#1)}
  \pgfmathsetmacro{\w}{min(\maxbarwidth, \maxbarwidth*\absval/\deltamax)}
  \fill[DeltaBlueNeg!75] (0,0) rectangle (\w cm,\barheight cm);
}}

\newcommand{\decDeep}[1]{%
{\scriptsize\textcolor{DeltaBlueNeg!90!black}{\,#1}}%
\,\tikz[baseline=-0.55ex]{
  \pgfmathsetmacro{\absval}{abs(#1)}
  \pgfmathsetmacro{\w}{min(\maxbarwidth, \maxbarwidth*\absval/\deltamax)}
  \fill[DeltaBlueNeg!90!black] (0,0) rectangle (\w cm,\barheight cm);
}}

\section{Experiments, Results, and Analysis}
\begin{table*}[t]
    \centering
    \scriptsize
    \renewcommand{\arraystretch}{1.18}
    \setlength{\tabcolsep}{2.4pt}
    \caption{\textbf{Main results on interactive benchmarks across different models.}
    For ALFWorld, we report the success rate for each task type and the overall result.
    Delta annotations indicate the absolute gain relative to the corresponding baseline:
    \textbf{ReAct} for prompting methods and \textbf{GRPO} for RL-based methods.
    The prompting/RL variant \textit{ReAct w/ \ourmethodshort{}} and \textit{GRPO w/ \ourmethodshort{}} is highlighted in light gray, and the full RL variant (\textit{GRPO w/ \ourmethodfullshort{}}) is highlighted in dark gray (\underline{SG = Self-Guidance, and GR = Guidance Reward}).
    For ALFWorld, delta annotations are shown only on the \textbf{All} column for readability.}
    \label{tab:main_results}
    \resizebox{\textwidth}{!}{
    \begin{tabular}{lllllllllllll}
        \toprule
        \multirow{2}{*}{Type} & \multirow{2}{*}{Method}
        & \multicolumn{7}{c}{ALFWorld}
        & \multicolumn{2}{c}{ScienceWorld}
        & \multicolumn{2}{c}{WebShop} \\
        \cmidrule(lr){3-9} \cmidrule(lr){10-11} \cmidrule(lr){12-13}
        &
        & Pick & Look & Clean & Heat & Cool & Pick2 & All
        & Succ. & Score
        & Succ. & Score \\
        \midrule

        % ------------------------ Qwen3 4B ------------------------
        \multicolumn{13}{l}{\textbf{Qwen3-4B}} \\
        \midrule
        \multirow{4}{*}{Prompting}
        & ReAct
        & 42.4 & 16.7 & 0.0 & 8.3 & 9.1 & 28 & 20.3 & 9.9 & 12.4 & 11.3 & 31.1 \\

        & Reflexion
        & 68.6 & 38.5 & 37.0 & 25.0 & 32.0 & 4.2 & 37.1 & 16.4 & 10.9 & 24.0 & 57.8 \\

        & ReFlAct
        & 74.3 & 38.5 & 25.9 & 25.0 & 32.0 & 16.7 & 38.6 & 9.0 & 11.8 & 16.0 & 35.1 \\

        & \hll \textbf{ReAct w/ SG}
        & \hll 87.8 & \hll 50.0 & \hll 29.2 & \hll 58.3 & \hll 27.3 & \hll 80.0 & \hll 58.6
        & \hll 10.4 & \hll 13.1 
        & \hll 17.6 & \hll 42.7 \\
        \cmidrule(lr){1-13}

        \multirow{3}{*}{RL Training}
        & GRPO
        & 95.7 & 88.9 & 100 & 92.3 & 67.7 & 86.4 & 86.7 & 59.3 & 51.4 & 71.9 & 84.3 \\

        & \hll GRPO w/ \ourmethodshort{}
        & \hll 97.1 & \hll 80.0 & \hll 96.2 & \hll 100 & \hll \textbf{87.5} & \hll 80.0 & \hll 91.4\incLite{4.7}
        & \hll 62.6\incLite{3.3}
        & \hll 60.4\incLite{9.0}
        & \hll 77.3\incLite{5.4}
        & \hll 87.0\incLite{2.7} \\

        & \hld \textbf{GRPO w/ \ourmethodfullshort{}}
        & \hld \textbf{100} & \hld \textbf{100} & \hld \textbf{100} & \hld \textbf{100} & \hld 87.0 & \hld \textbf{93.8} & \hld \textbf{96.9}\incDeep{10.2}
        & \hld \textbf{65.0}\incDeep{5.7} 
        & \hld \textbf{61.6}\incDeep{10.2}
        & \hld \textbf{78.1}\incDeep{6.2}
        & \hld \textbf{87.8}\incDeep{3.5} \\
        \midrule

        % ------------------------ Qwen3 1.7B ------------------------
        \multicolumn{13}{l}{\textbf{Qwen3-1.7B}} \\
        \midrule
        \multirow{4}{*}{Prompting}
        & ReAct
        & 24.2 & 12.5 & 0.0 & 0.0 & 0.0 & 12.1 & 10.2 & 0.9 & 1.0 & 4.7 & 41.6 \\

        & Reflexion
        & 22.9 & 15.4 & 14.8 & 18.8 & 16.0 & 8.3 & 16.4 & 12.3 & 11.8 & 16.0 & 39.8 \\

        & ReFlAct
        & 28.6 & 23.1 & 7.4 & 6.3 & 8.0 & 0.0 & 12.9 & 0.9 & 1.2 & 4.4 & 37.5 \\

        & \hll\textbf{ReAct w/ SG}
        & \hll54.2 & \hll14.3 & \hll18.5 & \hll0.0 & \hll12.1 & \hll19.0 & \hll21.1
        & \hll1.4 & \hll5.3 
        & \hll5.4 & \hll39.7 \\
        \cmidrule(lr){1-13}

        \multirow{3}{*}{RL Training}
        & GRPO
        & 93.1 & 100 & 83.3 & 52.6 & 54.5 & 60.7 & 72.7 & 42.6 & 23.5 & 32.0 & 63.0 \\

        & \hll GRPO w/ \ourmethodshort{}
        & \hll94.1 & \hll100 & \hll66.7 & \hll76.0 & \hll68.0 & \hll83.3 & \hll81.3\incLite{8.6}
        & \hll49.3\incLite{6.7} 
        & \hll25.3\incLite{1.8} 
        & \hll49.2\incLite{17.2}
        & \hll76.7\incLite{13.7} \\

        & \hld\textbf{GRPO w/ \ourmethodfullshort{}}
        & \hld\textbf{100} & \hld\textbf{100} & \hld\textbf{78.1} & \hld\textbf{100} & \hld\textbf{78.8} & \hld\textbf{100} & \hld\textbf{89.8}\incDeep{17.1}
        & \hld\textbf{52.1}\incDeep{9.5}
        & \hld\textbf{28.0}\incDeep{4.5}
        & \hld\textbf{56.3}\incDeep{24.3}
        & \hld\textbf{79.4}\incDeep{16.4} \\
        \midrule

        % ------------------------ Qwen2.5 7B Instruct ------------------------
        \multicolumn{13}{l}{\textbf{Qwen2.5-7B-Instruct}} \\
        \midrule
        \multirow{4}{*}{Prompting}
        & ReAct
        & 65.5 & 66.7 & 41.9 & 57.9 & 17.4 & 15.0 & 42.2 & 3.1 & 3.3 & 7.8 & 19.6 \\

        & Reflexion
        & 77.1 & 61.5 & 40.7 & 31.3 & 24.0 & 33.3 & 45.0 & 21.8 & 10.0 & 29.0 & 62.1 \\

        & ReFlAct
        & 74.3 & 46.2 & 51.9 & 50.0 & 28.0 & 33.3 & 47.1 & 8.5 & 6.0 & 10.2 & 28.6 \\

        & \hll\textbf{ReAct w/ SG}
        & \hll87.9 & \hll50.0 & \hll68.4 & \hll41.7 & \hll17.4 & \hll48.5 & \hll55.5
        & \hll10.0 & \hll11.4
        & \hll13.3 & \hll31.4 \\
        \cmidrule(lr){1-13}

        \multirow{3}{*}{RL Training}
        & GRPO
        & 100 & 72.7 & 100 & 88.9 & 58.8 & 89.5 & 83.6 & 64.0 & 62.7 & 65.6 & 77.5 \\

        & \hll GRPO w/ \ourmethodshort{}
        & \hll100 & \hll91.7 & \hll100 & \hll89.5 & \hll\textbf{100} & \hll70.8 & \hll92.2\incLite{8.6}
        & \hll72.5\incLite{8.5}
        & \hll69.4\incLite{6.7}
        & \hll76.6\incLite{11.0}
        & \hll89.3\incLite{11.8} \\

        & \hld\textbf{GRPO w/ \ourmethodfullshort{}}
        & \hld\textbf{100} & \hld\textbf{100} & \hld\textbf{100} & \hld\textbf{100} & \hld90.5 & \hld\textbf{84.0} & \hld\textbf{95.3}\incDeep{11.7}
        & \hld\textbf{75.4}\incDeep{11.4}
        & \hld\textbf{72.6}\incDeep{11.9}
        & \hld\textbf{80.5}\incDeep{14.9}
        & \hld\textbf{90.1}\incDeep{12.6} \\
        \bottomrule
    \end{tabular}
    }
\end{table*}
\label{sec:experiments}

\newcommand{\posdelta}[1]{\textcolor{ForestGreen}{\scriptsize (+#1)}}
\newcommand{\reflactposdelta}[1]{\textcolor{RoyalBlue}{\scriptsize (+#1)}}
\newcommand{\negdelta}[1]{\textcolor{BrickRed}{\scriptsize (-#1)}}

\newcommand{\pmstd}[1]{\textsuperscript{\scriptsize$\pm$#1}}

% We evaluate whether a rater can serve as an effective internal reward source for multi-turn language-agent training. Our experiments are designed to answer three questions. First, is a naive rater already effective before dedicated training, and how does its effect vary across environments of different complexity? Second, can a learned rater improve GRPO both as rollout guidance and as an internal reward source? Third, how important are online rater updates and staged reward activation to the final gains?

\subsection{Experimental Setup}
We evaluate on three benchmarks: \textbf{ALFWorld}~\citep{shridhar2021alfworldaligningtextembodied}, \textbf{ScienceWorld}~\citep{wang2022scienceworldagentsmarter5th}, and \textbf{WebShop}~\citep{yao2023webshopscalablerealworldweb}, covering a range of task complexity from structured household navigation to multi-step scientific reasoning and noisy web interaction. All experiments are conducted in the text-only setting, using \textbf{Qwen3-1.7B}, \textbf{Qwen3-4B}~\citep{yang2025qwen3technicalreport}, and \textbf{Qwen2.5-7B-Instruct}~\citep{qwen2025qwen25technicalreport} as backbone models.
 
We evaluate in two settings. In the \textbf{prompting setting}, we compare \textbf{ReAct}~\citep{yao2023reactsynergizingreasoningacting}, \textbf{Reflexion}~\citep{shinn2023reflexionlanguageagentsverbal}, \textbf{ReFlAct-style}~\citep{kim2025reflactworldgroundeddecisionmaking},\footnote{We re-implement this baseline following the paper description, as no official implementation was available at the time of writing.} and \textbf{ReAct w/ Self-Guidance}, which adds our self-guidance signal to ReAct without any RL training. In the \textbf{RL setting}, we compare \textbf{GRPO}, \textbf{\ourmethod{}} (self-guidance during rollout only), and \textbf{\ourmethodfull{}} (full method with self-guidance reward and staged schedule), isolating the contribution of each component. 
\paragraph{Training Setup.}
The self-guidance reward coefficient follows the trapezoid schedule 
described in Section~\ref{sec:schedule}: zero before step 40, 
linearly warmed up to 1 over steps 40--50, held at 1 over steps 
50--70, and annealed back toward 0 over steps 70--80, detailed in Appendix~\ref{app:experiment_details}.

\vspace{-0.1cm}
\subsection{Main Results}
\begin{figure}[t]
    \centering
    \includegraphics[width=\textwidth]{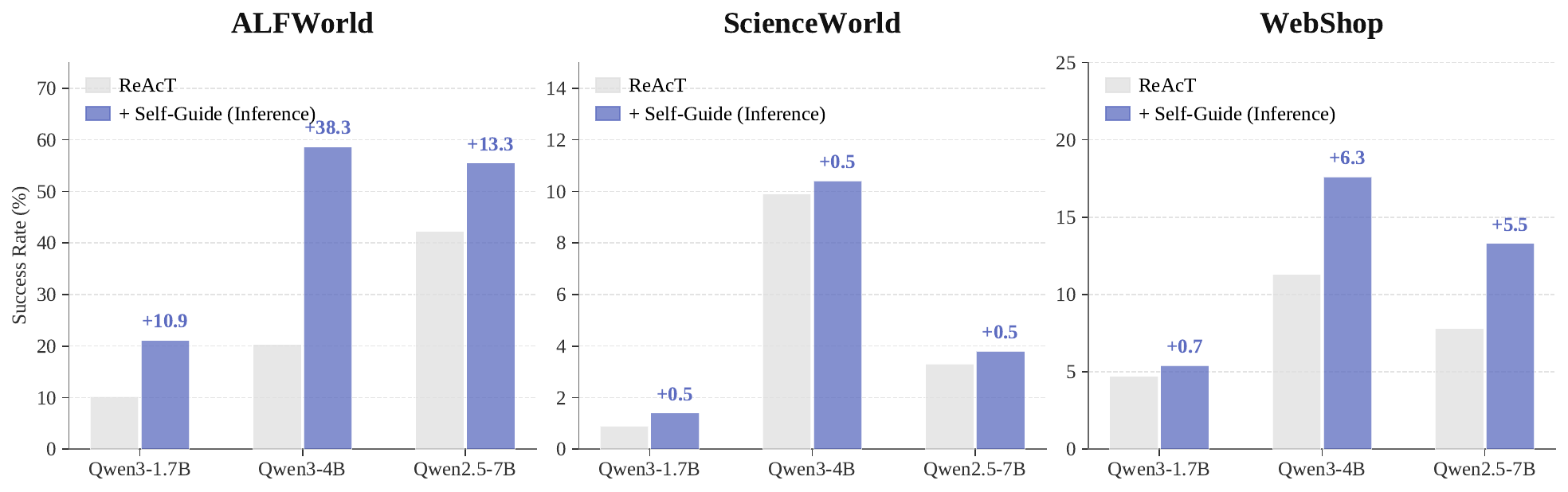}
    \caption{Self-guidance without training already improves performance in structured environments (ALFWorld) but yields inconsistent gains in more complex ones (WebShop), indicating that self-guidance quality depends on task familiarity.}
    \label{fig:motivation_naive_rater}
\end{figure}

\vspace{-0.1cm}
Table~\ref{tab:main_results} reports the main results across all environments and backbone models.
Looking first at the prompting setting, \textbf{ReAct w/ Self-Guidance} already outperforms ReAct and Reflexion on ALFWorld, where task progress is relatively structured and self-guidance signals are more reliable.
However, the same frozen self-guidance yields only marginal gains on WebShop, where intermediate decisions are more ambiguous and the model has limited prior understanding of the environment dynamics (Figure~\ref{fig:motivation_naive_rater}).
This contrast confirms that self-guidance quality depends on task familiarity, and motivates learning it jointly with the policy rather than treating it as a fixed component.
 
In the RL setting, \textbf{\ourmethodfull{}} consistently improves over the GRPO baseline across all three benchmarks and all backbone models.
\textbf{\ourmethod{}} (self-guidance during rollout only) already brings meaningful gains over GRPO, demonstrating that decision-time steering alone is beneficial.
Adding the self-guidance reward with the staged schedule (\textbf{\ourmethodfull{}}) brings further improvement, confirming that the internal guidance reward provides complementary signal beyond what decision-time guidance alone contributes.
\vspace{-0.1cm}
\paragraph{Error Analysis.}
Beyond success rate, self-guidance also improves the error distribution 
across models and tasks even without training, suggesting its benefits 
are not limited to the RL setting. We provide a detailed analysis in 
Appendix~\ref{app:error_analysis}.

\subsection{Self-Guidance Reward Learning Dynamics}
\begin{figure*}[t]
    \centering
    \begin{subfigure}[t]{0.32\textwidth}
        \centering
        \includegraphics[width=\textwidth]{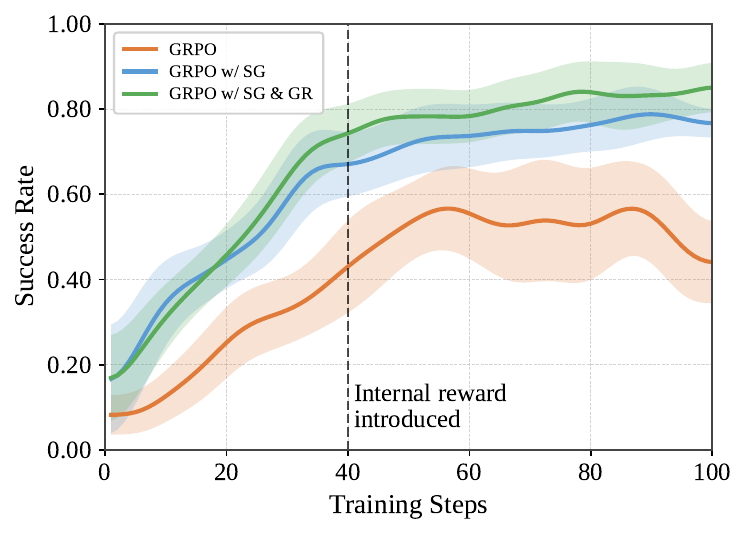}
        \caption{ALFWorld}
        \label{fig:train_curve_alfworld_qwen3_1p7b}
    \end{subfigure}
    \hfill
    \begin{subfigure}[t]{0.32\textwidth}
        \centering
        \includegraphics[width=\textwidth]{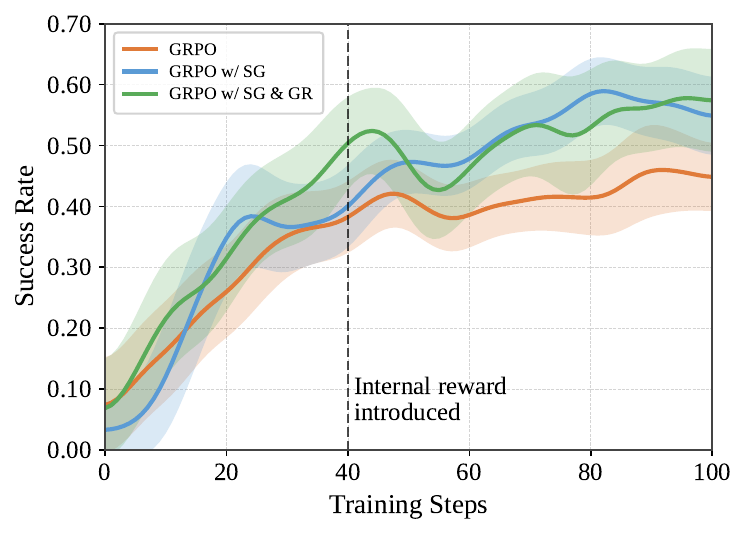}
        \caption{ScienceWorld}
        \label{fig:train_curve_scienceworld_qwen3_1p7b}
    \end{subfigure}
    \hfill
    \begin{subfigure}[t]{0.32\textwidth}
        \centering
        \includegraphics[width=\textwidth]{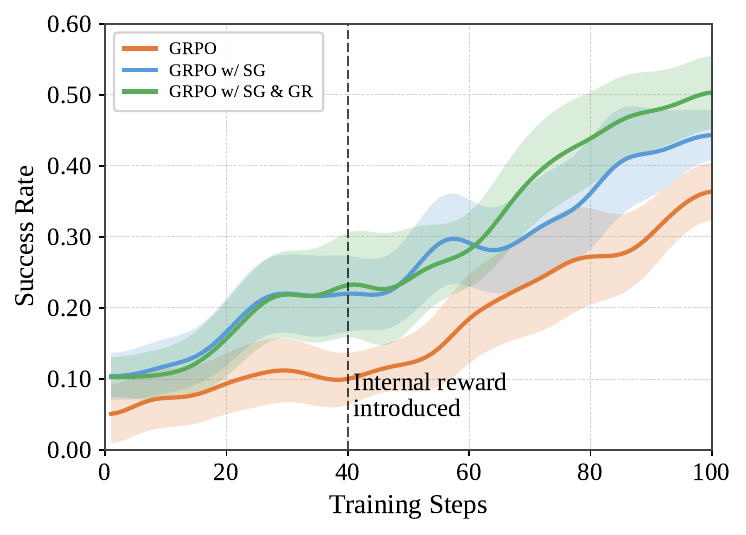}
        \caption{WebShop}
        \label{fig:train_curve_webshop_qwen3_1p7b}
    \end{subfigure}
    \caption{Training curves on the three environments with Qwen3-1.7B as base model.}
    \label{fig:qwen3_1p7b_training_curves}
\end{figure*}

Figure~\ref{fig:qwen3_1p7b_training_curves} shows the training dynamics 
across all three environments.
\textbf{GRPO w/ SG} outperforms vanilla GRPO from the very first steps, 
demonstrating that decision-time self-guidance provides immediate benefits 
independent of any reward signal.
At around step 40, when the staged schedule activates the self-guidance 
reward, \textbf{GRPO w/ SG \& GR} begins to separate from \textbf{GRPO w/ SG} 
and sustains a growing advantage through the remainder of training.
This divergence is the self-reinforcing loop in action: once self-guidance 
quality has matured through acting, promoting it to a reward source 
accelerates policy improvement in a way that guidance alone cannot.

\textbf{Compatibility with Other RL Algorithms}: Since our method only modifies the training signal while leaving the 
underlying optimizer unchanged, it is compatible with other RL algorithms. 
We verify this with DAPO, where the same gains are observed 
(Appendix~\ref{sec:ablation_optimizer}).
\subsection{Ablation on Stage-Wise Guidance Reward Scheduling}
\label{sec:ablation_schedule_timing}
% Our method uses a staged schedule to progressively activate the guidance reward during training. We now ask whether such scheduling is necessary, or whether one can simply enable guidance reward earlier or keep it active throughout training.
% We compare the following variants: \textbf{Vanilla GRPO}, which uses no self-guidance or guidance reward at all; \textbf{Self-guidance Only}, where self-guidance is used during rollout but provides no guidance reward; \textbf{Early Entry}, where guidance reward is introduced earlier than in our default schedule (at step 15 or step 25); \textbf{No Annealing}, where guidance reward is warmed up and reaches full strength but is not reduced later; and \textbf{GRPO w/ \ourmethodfullshort{}}, which uses our complete stage-wise schedule with rollout-only, warmup, full-internal-reward, and late-annealing phases.

\begin{figure}[t]
    \centering
    \begin{subfigure}[c]{0.48\linewidth}
        \centering
        \includegraphics[width=\linewidth]{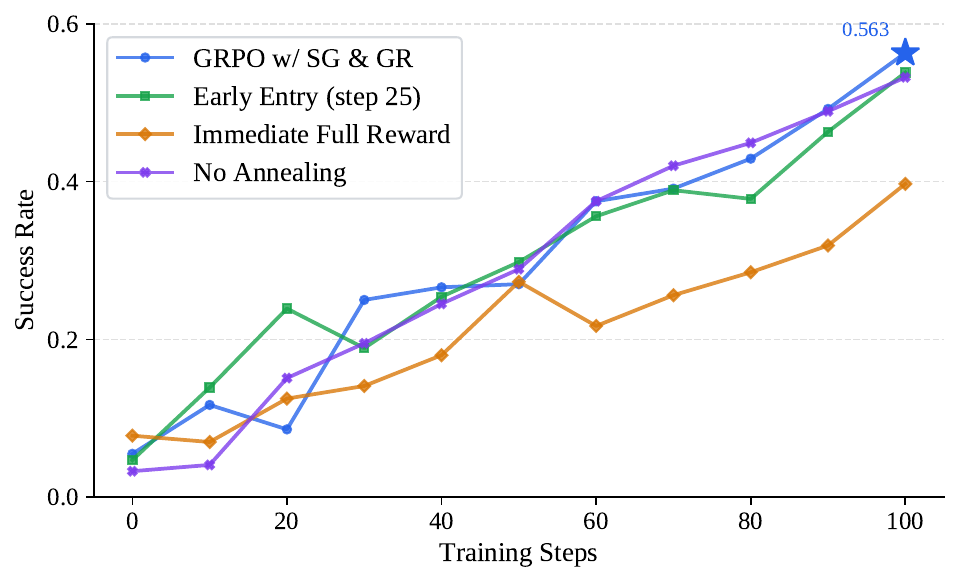}
    \end{subfigure}
    \hspace{0.01\linewidth}
    \begin{subfigure}[c]{0.45\linewidth}
        \centering
        \vspace{-1.6em}
        \small
        \renewcommand{\arraystretch}{1.1}
        \begin{tabular}{lc}
            \toprule
            Variant & Success Rate \\
            \midrule
            Vanilla GRPO & 32.0 \\
            Self-guidance Only & 49.2 \\
            Immediate Full Reward & 39.7 \\
            Early Entry (step 15) & 49.4 \\
            Early Entry (step 25) & \underline{53.8} \\
            No Annealing & 53.2 \\
            GRPO w/ \ourmethodfullshort{} & \textbf{56.3} \\
            \bottomrule
        \end{tabular}
    \end{subfigure}
    \caption{Ablation on stage-wise guidance-reward scheduling. \textbf{Left:} Training curves under different reward schedules. \textbf{Right:} Final validation success rates.}
    \label{fig:stage_ablation}
\end{figure}

% As shown in Figure~\ref{fig:stage_ablation}, directly introducing guidance reward too early underperforms the staged alternative, indicating that early self-guidance judgments are not yet reliable enough to serve as strong optimization targets. Removing late annealing also slightly degrades performance, suggesting that sustained full-strength guidance reward can eventually become overly restrictive. Overall, the best results are achieved by the full schedule, which balances early exploration, gradual reward activation, and late-stage relaxation.
We ablate the staged trust schedule by varying when and whether the self-guidance reward is introduced.
As shown in Figure~\ref{fig:stage_ablation}, introducing the guidance reward immediately (\textbf{Immediate Full Reward}: 39.7) performs worse than not using it at all (\textbf{Self-guidance Only}: 49.2), confirming that early self-guidance signals are not yet reliable enough to serve as optimization targets.
Entering earlier than our default (\textbf{Early Entry} at step 15 or 25) also underperforms, with gains increasing as the entry point moves later.
Removing late annealing (\textbf{No Annealing}: 53.2) slightly degrades performance relative to the full schedule (56.3), suggesting that sustained full-strength guidance reward eventually over-constrains the policy.
Together, these results validate the overall structure of the trapezoid schedule: a warm-up period before reward is introduced, and a reduction in reward weight before convergence matter more than the precise timing of each transition.

\subsection{Ablation on Co-Evolution}
\label{sec:offline_vs_online_rater}
A natural alternative is to first distill a stronger self-guidance signal 
from a larger teacher model offline, then use it to bootstrap downstream 
RL training. We test this by distilling a Qwen3-1.7B model from Qwen3-32B, 
then plugging the result into our full \ourmethodfull{} framework.

\begin{figure*}[t]
    \centering
    \includegraphics[width=\textwidth]{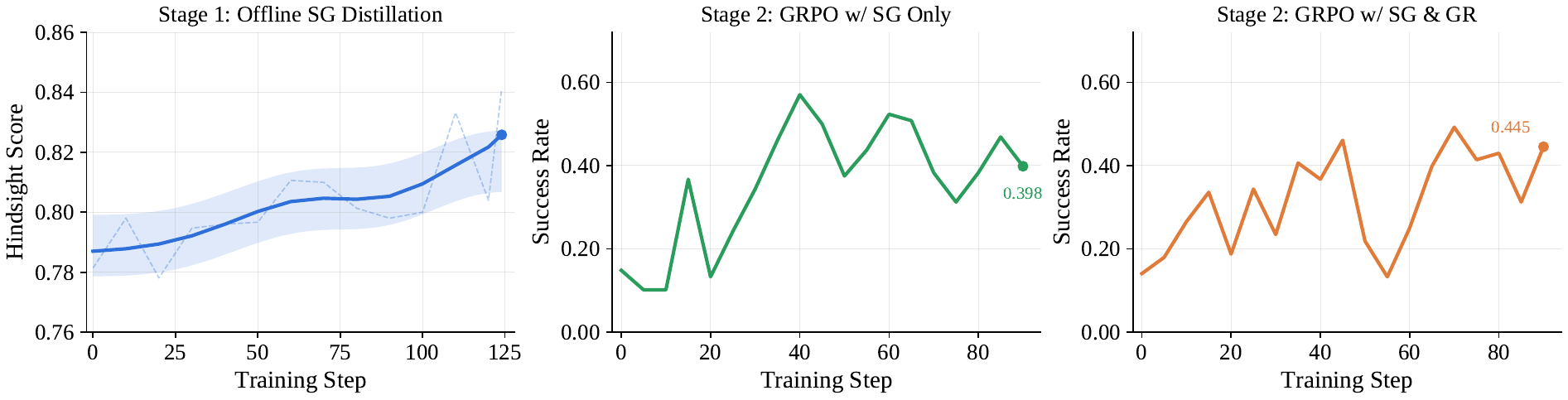}
    \caption{
\textbf{Offline self-guidance distillation versus online co-evolution.}
Although the student improves under offline distillation (a), the distilled 
model does not transfer reliably to downstream RL: using it as guidance 
reward (b) or for inference-time guidance only (c) yields unstable or 
unsustained gains, supporting the need for online co-evolution.
}
    \label{fig:offline_sg_main}
\end{figure*}

As shown in Figure~\ref{fig:offline_sg_main}, the distilled self-guidance 
does not transfer reliably to downstream RL: using it as a dense guidance 
reward leads to unstable training, and using it for rollout guidance only 
also fails to yield sustained gains.
We attribute this to a distribution mismatch: a model trained on fixed 
offline trajectories is poorly calibrated to the evolving rollout 
distribution of the online policy.
Our online approach avoids this by co-evolving self-guidance and policy 
throughout training. We further verify that co-evolution strengthens the 
self-guidance model itself: fixing the policy and varying only the 
self-guidance checkpoint shows consistent gains as training progresses
(Appendix~\ref{app:coevolving_rater}).
% \begin{figure*}[t]
%     \centering
%     \begin{subfigure}[b]{0.32\textwidth}
%         \centering
%         \includegraphics[width=\textwidth]{figures/figure_6_combined.pdf}
%         \caption{The student self-guidance model improves steadily under the offline distillation objective.}
%         \label{fig:offline_rater_stage1_main}
%     \end{subfigure}
%     \hfill
%     \begin{subfigure}[b]{0.32\textwidth}
%         \centering
%         \includegraphics[width=\textwidth]{figures/stage2_rater_reward.png}
%         \caption{Using the distilled self-guidance model as dense internal reward leads to unstable downstream RL.}
%         \label{fig:offline_rater_stage2_reward_main}
%     \end{subfigure}
%     \hfill
%     \begin{subfigure}[b]{0.32\textwidth}
%         \centering
%         \includegraphics[width=\textwidth]{figures/stage2_rollout_only.png}
%         \caption{Using the distilled model only for self-guidance also fails to produce sustained gains.}
%         \label{fig:offline_rater_stage2_rollout_main}
%     \end{subfigure}
%     \caption{Comparison between offline self-guidance distillation and self-guidance-policy co-evolution. Although the student self-guidance model improves under the offline proxy objective, the distilled model does not transfer reliably to downstream agent RL.}
%     \label{fig:offline_rater_main}
% \end{figure*}

% \subsection{Analysis}

% To better understand why the proposed method works, we analyze the resulting agent behavior at the trajectory level. If rater-based internal reward indeed improves credit assignment, it should not only increase final success rate, but also improve the quality of intermediate decision making.

% We therefore examine process-level statistics such as trajectory efficiency, redundant actions, and cumulative rater reward on successful and failed trajectories. Table~\ref{tab:analysis} provides a placeholder summary of these analyses. In general, we expect our method to produce more efficient trajectories and a clearer separation between successful and failed rollouts in terms of accumulated internal reward.

% Finally, Figure~\ref{fig:rater_analysis} can be used to visualize how the learned rater behaves, for example by comparing cumulative rater reward between successful and failed trajectories, or by showing the distribution of positive/neutral/negative ratings over training. Such analysis helps verify that the rater is not merely adding noise, but is learning a meaningful internal signal aligned with long-horizon progress.

\section{Conclusion}

We proposed Self-Guide, a unified framework that co-evolves policy and internal reward within a single language agent. By having the agent generate verbal self-guidance at each decision step and repurposing the same signal as dense internal reward for training, our method creates a self-reinforcing loop that requires no external reward model or separate evaluator. A stage-wise trust schedule stabilizes this co-evolution by activating and later annealing the internal reward as the guidance signal matures. Experiments on ALFWorld, ScienceWorld, and WebShop show that decision-time self-guidance alone already yields meaningful improvements and that jointly co-evolving it with internal reward further advances performance over strong GRPO baselines across multiple model scales.

\section*{Ethics Statement}
This work studies learned internal guidance rewards for language agents in interactive benchmark environments. Our experiments exclusively use publicly available benchmark environments, and do not involve non-public data or personally identifiable information.

\section*{LLM Usage}
We make use of GPT-5.4 and Claude Opus 4.6  to check potential grammar mistakes and refine our writing. We adopt GPT-4.1 to help conduct our experiments.

\bibliography{colm2026_conference}
\bibliographystyle{colm2026_conference}

\appendix
% ============================================================
%  Appendix: Experiment Details
%  Requires: tcolorbox, enumitem, amsmath
% ============================================================

\section{Experiment Details}
\label{app:experiment_details}

We describe the prompts and hyperparameters used in each environment.
All RL experiments are implemented using the \texttt{verl-agent}~\citep{feng2025groupingrouppolicyoptimizationllm} framework.
Self-guidance prompt (Figure~\ref{fig:rater_prompt}) is identical across all three environments.
Environment-specific actor prompts and training configurations follow.

% -------------------------------------------------------
%  RATER PROMPT  (shared, violet)
% -------------------------------------------------------

% -------------------------------------------------------
%  ALFWORLD  (blue)
% -------------------------------------------------------
\subsection*{ALFWorld}

\paragraph{Hyperparameters.}
The maximum prompt length is \texttt{4096} tokens and the maximum response length is \texttt{1024} tokens.
Each episode allows up to \texttt{50} environment steps. The rollout temperature is \texttt{1.0} and the validation temperature is \texttt{0.4}. The total number of training steps is \texttt{100}.
\begin{figure*}[t]
\centering
\begin{tcolorbox}[
    colback=violet!5,
    colframe=violet!55!black,
    boxrule=0.8pt,
    arc=4mm,
    width=0.95\textwidth,
    title={Self-Guidance Prompt (Shared Across All Environments)},
    fonttitle=\bfseries
]
\small
\textbf{TASK=rate}\\[2pt]
You are self-guidance for action-observation trajectories.

\medskip
\noindent\textbf{INPUT}\\
Initial observation \& task:\\[2pt]
\texttt{\{initial\_obs\}}

\smallskip
\noindent Current trajectory (steps $1\,..\,t$):\\[2pt]
\texttt{\{sub\_traj\}}

\medskip
\noindent\textbf{YOUR TASK}\\
Using the sub-trajectory in the context of the initial observation, assign a label in
\{\texttt{negative}, \texttt{neutral}, \texttt{positive}\} to step $t$, and provide a short reasoning on why this label makes sense.

\medskip
\noindent\textbf{LABEL MEANINGS}

\noindent\hspace{1em}\textbullet\ \textbf{positive} --- making progress: clearly moves closer to task completion or gathers necessary information later used to advance.\\[2pt]
\noindent\hspace{1em}\textbullet\ \textbf{neutral} --- not clearly helpful or harmful.\\[2pt]
\noindent\hspace{1em}\textbullet\ \textbf{negative} --- unproductive or harmful (wasted steps, regressions, contradictions, stuck).

\medskip
\noindent\textbf{OUTPUT FORMAT} (exactly \textbf{two} lines, no extra text):\\[4pt]
\texttt{Reason: <brief explanation>}\\
\texttt{Label: negative|neutral|positive}

\medskip
\noindent Now output the two lines:
\end{tcolorbox}
\caption{Self-guidance prompt template, applied identically across ALFWorld, SciWorld, and WebShop.}
\label{fig:rater_prompt}
\end{figure*}
\begin{figure*}[t]
\centering
\begin{tcolorbox}[
    colback=blue!4,
    colframe=blue!55!black,
    boxrule=0.8pt,
    arc=4mm,
    width=0.95\textwidth,
    title={Actor Prompt Template for ALFWorld},
    fonttitle=\bfseries
]
\small
Interact with a household to solve a task.

\smallskip
\noindent Respond with exactly one JSON object on a single line using this format:

\smallskip
\noindent\texttt{\{}\\\
\hspace*{1.5em}\texttt{"think": "optional short reasoning (can be empty)",}\\
\hspace*{1.5em}\texttt{"action": "next action to take (required)"}\\
\texttt{\}}

\noindent\hspace{1em}\textbullet\ \texttt{"action"} is required every turn.\\[2pt]
\noindent\hspace{1em}\textbullet\ \texttt{"think"} is optional; leave it empty if you do not need to reason explicitly.\\[2pt]
\noindent\hspace{1em}\textbullet\ Do not add any extra text outside the JSON.

\medskip
\noindent Here is your current action-observation history:

\smallskip
\noindent\texttt{\{trajectory\}}

\medskip
\noindent Intuition score for the CURRENT trajectory prefix: \texttt{\{self\_guidance\_score\}}\\
\texttt{\{reasoning\}}

\smallskip
\noindent\texttt{\{admissible\_actions\}}

\medskip
\noindent Choose the next action from the admissible actions using the self-guidance and reason.
Respond in the specified JSON format:
\end{tcolorbox}
\caption{Actor prompt template for ALFWorld. Admissible actions and the self-guidance are injected at each step.}
\label{fig:alfworld_prompt}
\end{figure*}

% -------------------------------------------------------
%  SCIWORLD  (green)
% -------------------------------------------------------
\subsection*{SciWorld}

\paragraph{Hyperparameters.}
The maximum prompt length is \texttt{4096} tokens and the maximum response length is \texttt{1024} tokens.
Each episode allows up to \texttt{30} environment steps. The rollout temperature is \texttt{1.0} and the validation temperature is \texttt{0.4}. The total number of training steps is \texttt{100}.

\paragraph{Note on action space.}
SciWorld can expose more than 100 admissible actions per step, making it impractical to enumerate them in every prompt.
Instead, we inject only the \emph{action format grammar} (i.e., the set of valid command templates) directly into the prompt and allow the agent to generate free-form actions that conform to this grammar.

\begin{figure*}[t]
\centering
\begin{tcolorbox}[
    colback=green!4,
    colframe=green!50!black,
    boxrule=0.8pt,
    arc=4mm,
    width=0.95\textwidth,
    title={Actor Prompt Template for SciWorld},
    fonttitle=\bfseries
]
\small
Interact with the ScienceWorld environment to solve the task.
In the environment, there are several rooms: kitchen, foundry, workshop, bathroom, outside, living room, bedroom, greenhouse, art studio, hallway.
You should explore the environment and find the items you need to complete the experiment.
You can teleport to any room in one step.

\medskip
\noindent\textbf{Available actions:}

\noindent\hspace{1em}\textbullet\ \texttt{open/close <OBJ>} --- open or close a container\\[1pt]
\noindent\hspace{1em}\textbullet\ \texttt{de/activate <OBJ>} --- activate or deactivate a device\\[1pt]
\noindent\hspace{1em}\textbullet\ \texttt{connect <OBJ> to <OBJ>} --- connect electrical components\\[1pt]
\noindent\hspace{1em}\textbullet\ \texttt{disconnect <OBJ>} --- disconnect electrical components\\[1pt]
\noindent\hspace{1em}\textbullet\ \texttt{use <OBJ> [on <OBJ>]} --- use a device/item (optionally on a target object)\\[1pt]
\noindent\hspace{1em}\textbullet\ \texttt{look around} / \texttt{look at <OBJ>} / \texttt{look in <OBJ>} --- inspect the environment\\[1pt]
\noindent\hspace{1em}\textbullet\ \texttt{read <OBJ>} --- read a note or a book\\[1pt]
\noindent\hspace{1em}\textbullet\ \texttt{move <OBJ> to <OBJ>} --- move an object to a container\\[1pt]
\noindent\hspace{1em}\textbullet\ \texttt{pick up <OBJ>} --- move an object to the inventory\\[1pt]
\noindent\hspace{1em}\textbullet\ \texttt{put down <OBJ>} --- drop an inventory item\\[1pt]
\noindent\hspace{1em}\textbullet\ \texttt{pour <OBJ> into <OBJ>} --- pour a liquid into a container\\[1pt]
\noindent\hspace{1em}\textbullet\ \texttt{dunk <OBJ> into <OBJ>} --- dunk a container into a liquid\\[1pt]
\noindent\hspace{1em}\textbullet\ \texttt{mix <OBJ>} --- chemically mix a container\\[1pt]
\noindent\hspace{1em}\textbullet\ \texttt{teleport to <LOC>} --- move to a new location\\[1pt]
\noindent\hspace{1em}\textbullet\ \texttt{eat <OBJ>} --- eat a food item\\[1pt]
\noindent\hspace{1em}\textbullet\ \texttt{flush <OBJ>} --- flush a toilet\\[1pt]
\noindent\hspace{1em}\textbullet\ \texttt{focus on <OBJ>} --- signal intent on a task object\\[1pt]
\noindent\hspace{1em}\textbullet\ \texttt{wait} / \texttt{wait1} --- take no action for 10/1 iteration(s)\\[1pt]
\noindent\hspace{1em}\textbullet\ \texttt{task} --- describe the current task\\[1pt]
\noindent\hspace{1em}\textbullet\ \texttt{inventory} --- list your inventory\\[1pt]
\noindent\hspace{1em}\textbullet\ \texttt{done} --- indicate that the task is complete

\medskip
\noindent Respond with exactly one JSON object using this format:

\smallskip
\noindent\texttt{\{}\\
\hspace*{1.5em}\texttt{"think": "optional short reasoning (can be empty)",}\\
\hspace*{1.5em}\texttt{"action": "next action to take (required)"}\\
\texttt{\}}

\noindent\hspace{1em}\textbullet\ \texttt{"action"} is required every turn.\\[2pt]
\noindent\hspace{1em}\textbullet\ \texttt{"think"} is optional; leave it empty if you do not need to reason explicitly.\\[2pt]
\noindent\hspace{1em}\textbullet\ Do not add any extra text outside the JSON.

\medskip
\noindent Here is your current action-observation history:

\smallskip
\noindent\texttt{\{trajectory\}}

\medskip
\noindent Intuition score for the CURRENT trajectory prefix: \texttt{\{self\_guidance\_score\}}\\
\texttt{\{reasoning\}}

\medskip
\noindent Choose the next action. Respond in the specified JSON format:
\end{tcolorbox}
\caption{Actor prompt template for SciWorld. Because the number of admissible actions can exceed 100 per step, the prompt enumerates only the action format grammar rather than the full per-step action list.}
\label{fig:sciworld_prompt}
\end{figure*}

% -------------------------------------------------------
%  WEBSHOP  (orange)
% -------------------------------------------------------
\subsection*{WebShop}

\paragraph{Hyperparameters.}
The maximum prompt length is \texttt{8192} tokens and the maximum response length is \texttt{1024} tokens.
Each episode allows up to \texttt{15} environment steps. The rollout temperature is \texttt{1.0} and the validation temperature is \texttt{0.4}. The total number of training steps is \texttt{100}.

\paragraph{Prompt design.}
The WebShop actor prompt follows the template introduced in GiGPO~\citep{feng2025groupingrouppolicyoptimizationllm}.

\begin{figure*}[t]
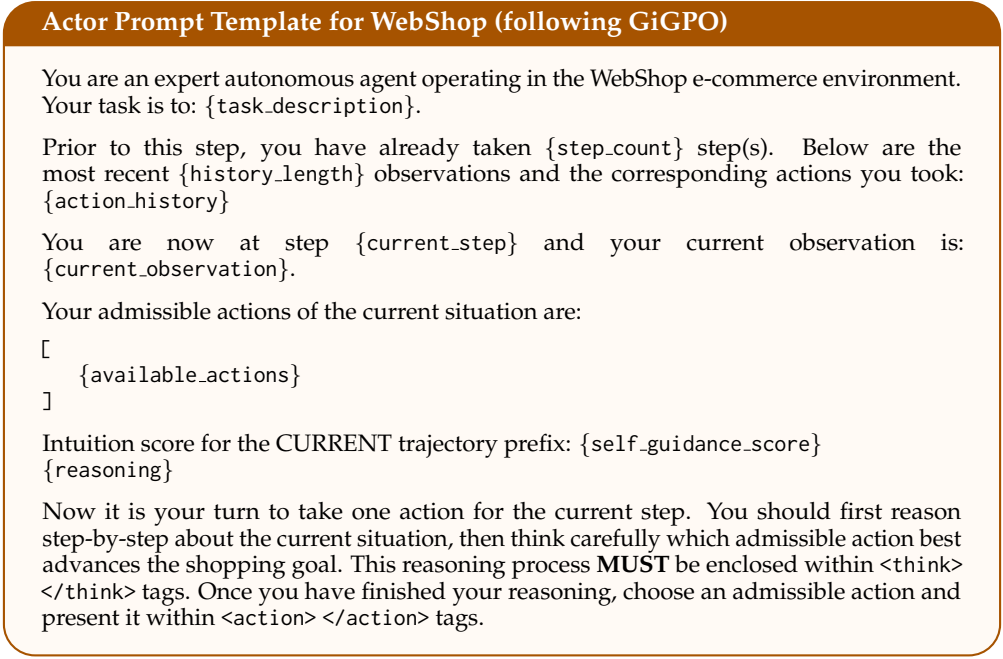

    \centering
    \begin{tcolorbox}[
        colback=orange!4,
        colframe=orange!65!black,
        boxrule=0.8pt,
        arc=4mm,
        width=0.95\textwidth,
        title={Actor Prompt Template for WebShop (following GiGPO)},
        fonttitle=\bfseries
    ]
        \small
        You are an expert autonomous agent operating in the WebShop e-commerce environment.\\
        Your task is to: \texttt{\{task\_description\}}.
        
        \medskip
        \noindent Prior to this step, you have already taken \texttt{\{step\_count\}} step(s).
        Below are the most recent \texttt{\{history\_length\}} observations and the corresponding actions you took:
        \texttt{\{action\_history\}}
        
        \medskip
        \noindent You are now at step \texttt{\{current\_step\}} and your current observation is:
        \texttt{\{current\_observation\}}.
        
        \medskip
        \noindent Your admissible actions of the current situation are:\\[4pt]
        \texttt{[}\\
        \hspace*{1.5em}\texttt{\{available\_actions\}}\\
        \texttt{]}
        
        \medskip
        \noindent Intuition score for the CURRENT trajectory prefix: \texttt{\{self\_guidance\_score\}}\\
        \texttt{\{reasoning\}}
        
        \medskip
        \noindent Now it is your turn to take one action for the current step.
        You should first reason step-by-step about the current situation, then think carefully which admissible action best advances the shopping goal.
        This reasoning process \textbf{MUST} be enclosed within \texttt{<think>} \texttt{</think>} tags.
        Once you have finished your reasoning, choose an admissible action and present it within \texttt{<action>} \texttt{</action>} tags.
    \end{tcolorbox}
    \caption{Actor prompt template for WebShop, adapted from GiGPO~\citep{feng2025groupingrouppolicyoptimizationllm}. Admissible actions and the self-guidance score are injected at each step.}
    \label{fig:webshop_prompt}
\end{figure*}

\section{Error Pattern Analysis of Trajectories}
\label{app:error_analysis}
We analyze failure modes across all three benchmarks to understand how self-guidance changes agent behavior beyond aggregate success rates. The analysis is conducted across the three benchmarks in the untrained setting using all three backbones evaluated in this work, namely \texttt{Qwen3-4B}, \texttt{Qwen3-1.7B}, and \texttt{Qwen2.5-7B-Instruct}, where self-guidance provides step-level guidance during rollout without any RL training.
 
\paragraph{Error Taxonomy.}
We identify three conceptual error categories, instantiated differently per environment based on its action space and task structure.
Table~\ref{tab:error_taxonomy_nav} and~\ref{tab:error_taxonomy_webshop} summarize the classification criteria. The same criteria are applied to both ReAct and ReAct+Self-Guidance trajectories.
 
\begin{table}[h]
    \centering
    \small
    \begin{tabular}{ll}
    \toprule
    \textbf{Category} & \textbf{Criterion (ALFWorld \& ScienceWorld)} \\
    \midrule
    \textbf{Looping}
      & Any action repeated $\geq\!5$ consecutively, or single action $\geq\!50\%$ of all steps \\[4pt]
    \textbf{Redundant Exploration}
      & $\geq\!8$ redundant \texttt{go to} revisits, or same object examined $\geq\!5$ times \\[4pt]
    \textbf{Wrong-Object Focus}
      & Irrelevant object examined $\geq\!4$ times, or $\geq\!7$ pick-ups of non-target objects \\
    \bottomrule
    \end{tabular}
    \caption{Error classification criteria for ALFWorld and ScienceWorld.}
    \label{tab:error_taxonomy_nav}
\end{table}

\begin{table}[h]
    \centering
    \small
    \begin{tabular}{ll}
    \toprule
    \textbf{Category} & \textbf{Criterion (WebShop)} \\
    \midrule
    \textbf{Query Looping}
      & Same search query issued $\geq\!3$ times (exact or normalized) \\[4pt]
    \textbf{Navigation Cycling}
      & \texttt{back to search} clicked $\geq\!3$ times without completing a purchase \\[4pt]
    \textbf{Premature Purchase}
      & Agent clicks \texttt{buy now} without selecting any required attribute \\
    \bottomrule
    \end{tabular}
    \caption{Error classification criteria for WebShop.}
    \label{tab:error_taxonomy_webshop}
\end{table}
 
\begin{figure}[h]
    \centering
    % Row 1: Model A
    \begin{subfigure}[b]{0.32\textwidth}
        \centering
        \includegraphics[width=\textwidth]{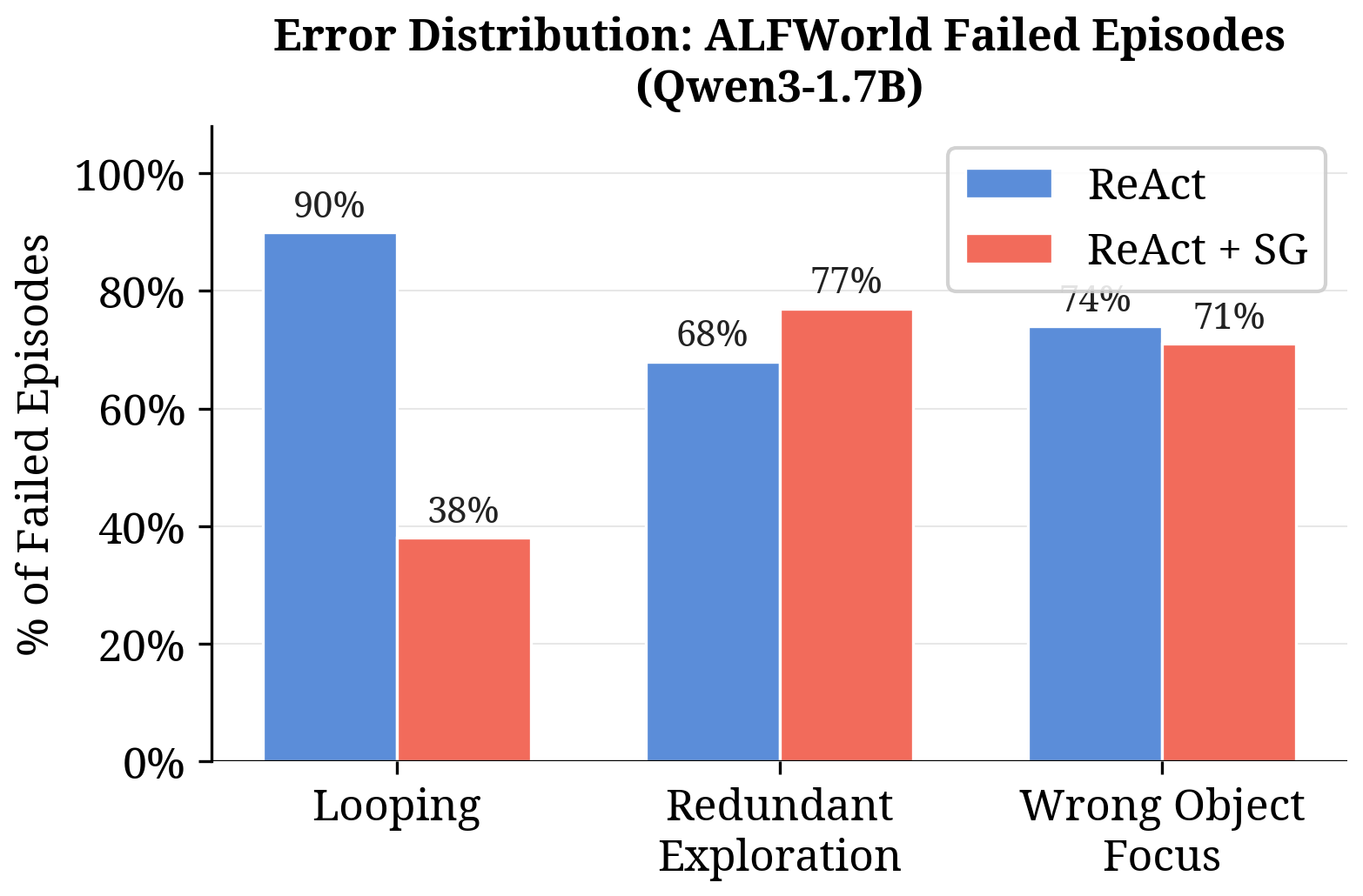}
    \end{subfigure}
    \hfill
    \begin{subfigure}[b]{0.32\textwidth}
        \centering
        \includegraphics[width=\textwidth]{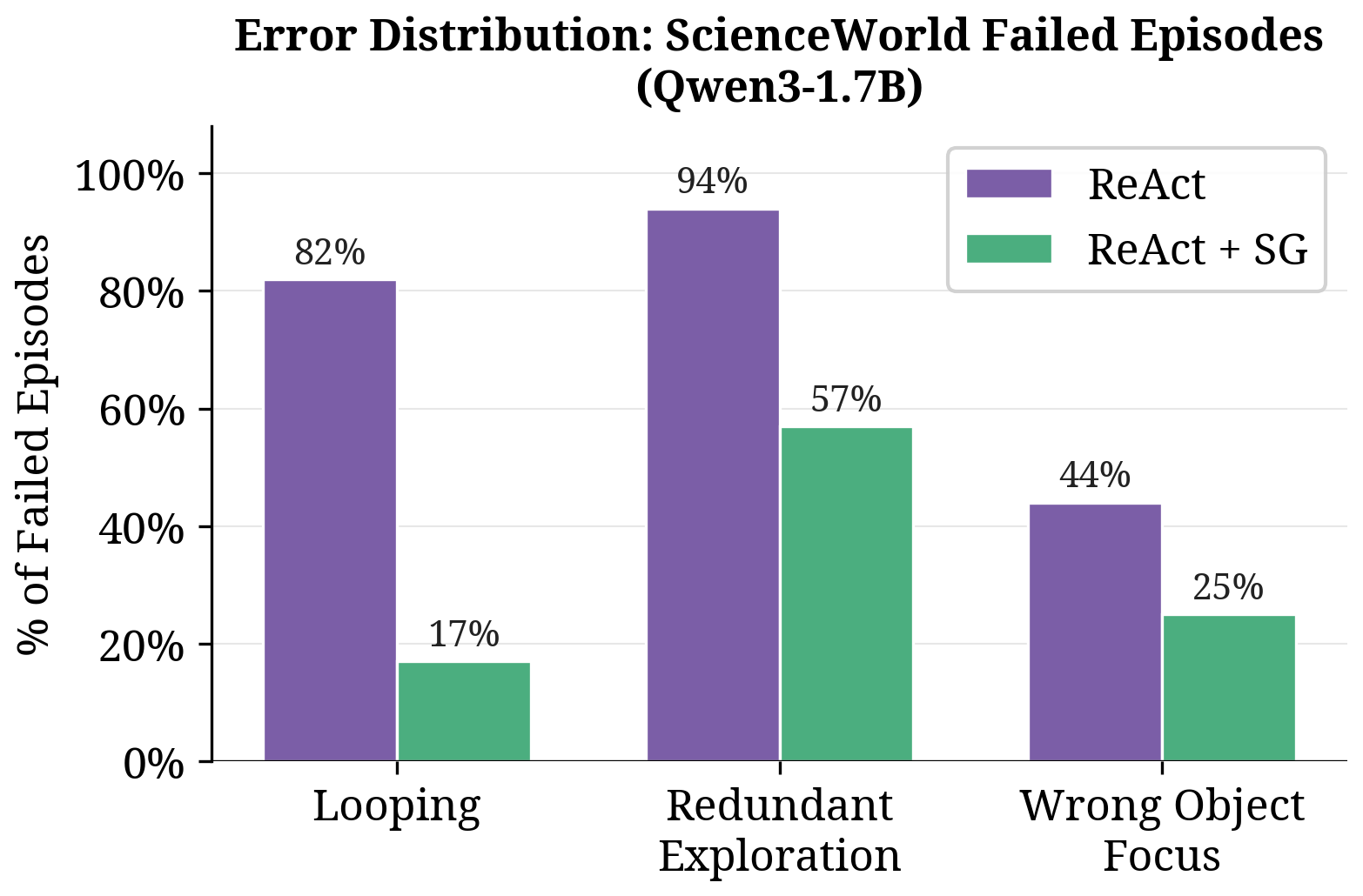}
    \end{subfigure}
    \hfill
    \begin{subfigure}[b]{0.32\textwidth}
        \centering
        \includegraphics[width=\textwidth]{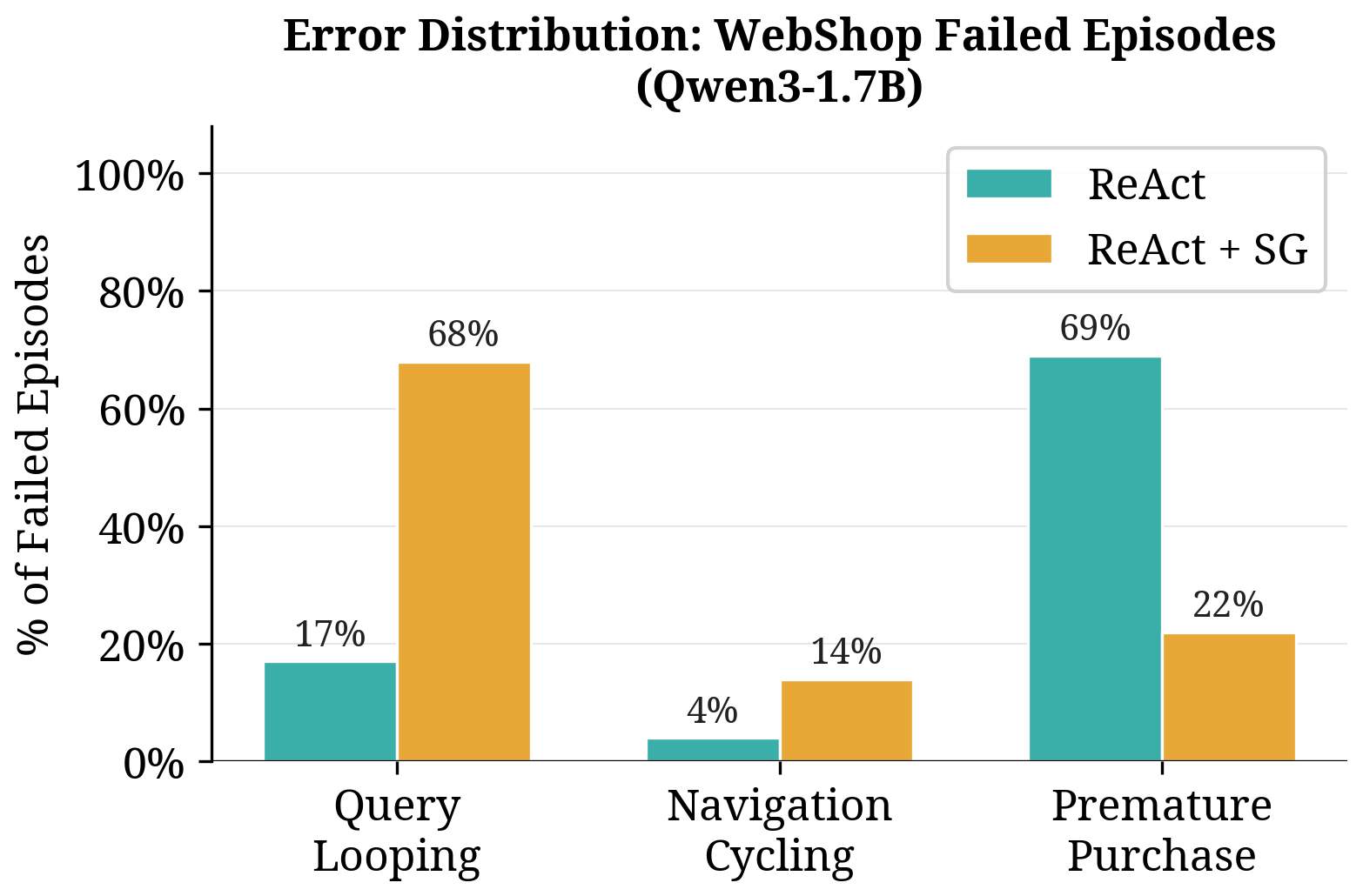}
    \end{subfigure}

    \vspace{0.5em}

    % Row 2: Model B
    \begin{subfigure}[b]{0.32\textwidth}
        \centering
        \includegraphics[width=\textwidth]{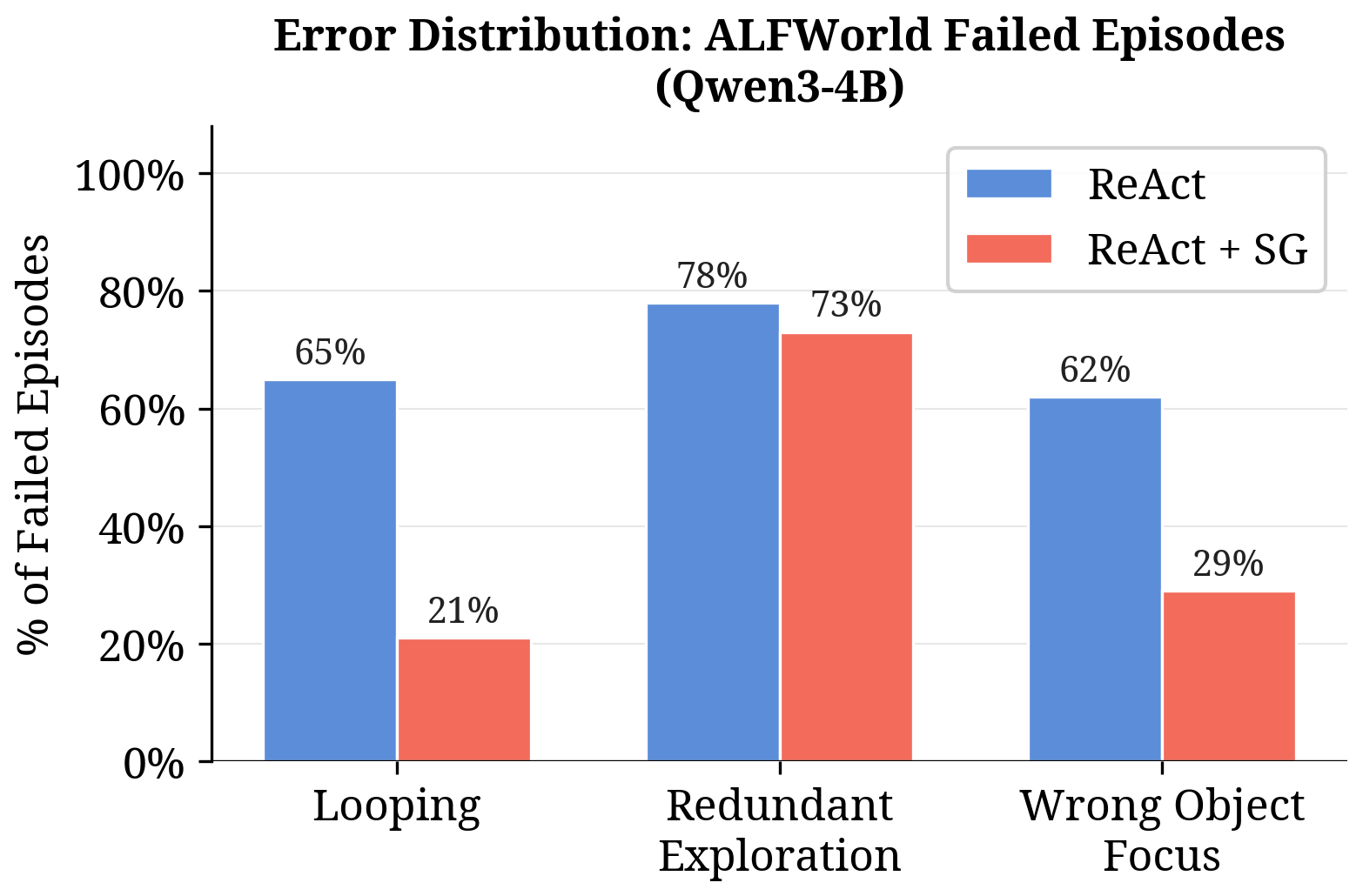}
        \label{fig:error_alfworld}
    \end{subfigure}
    \hfill
    \begin{subfigure}[b]{0.32\textwidth}
        \centering
        \includegraphics[width=\textwidth]{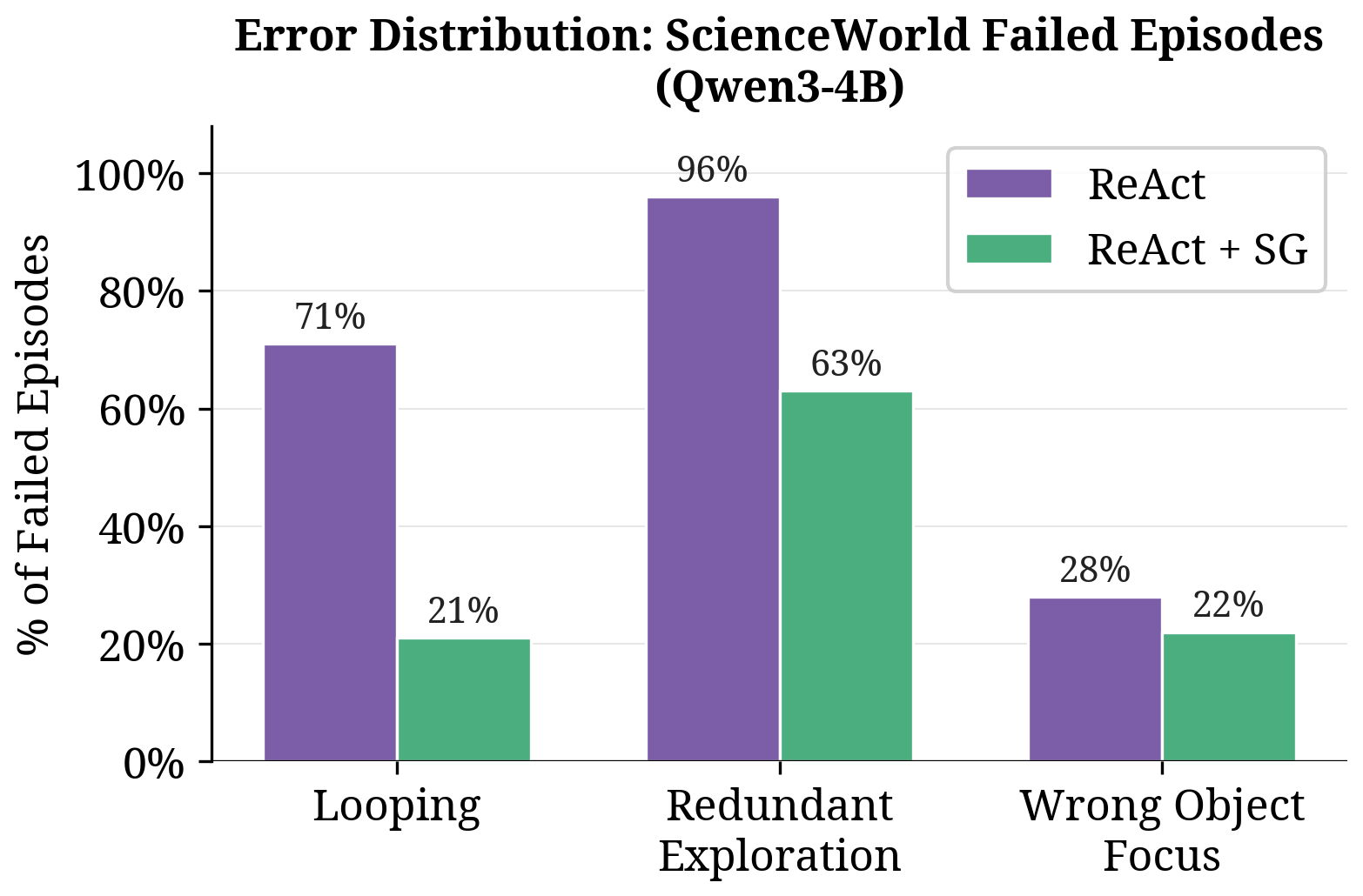}
        \label{fig:error_sciworld}
    \end{subfigure}
    \hfill
    \begin{subfigure}[b]{0.32\textwidth}
        \centering
        \includegraphics[width=\textwidth]{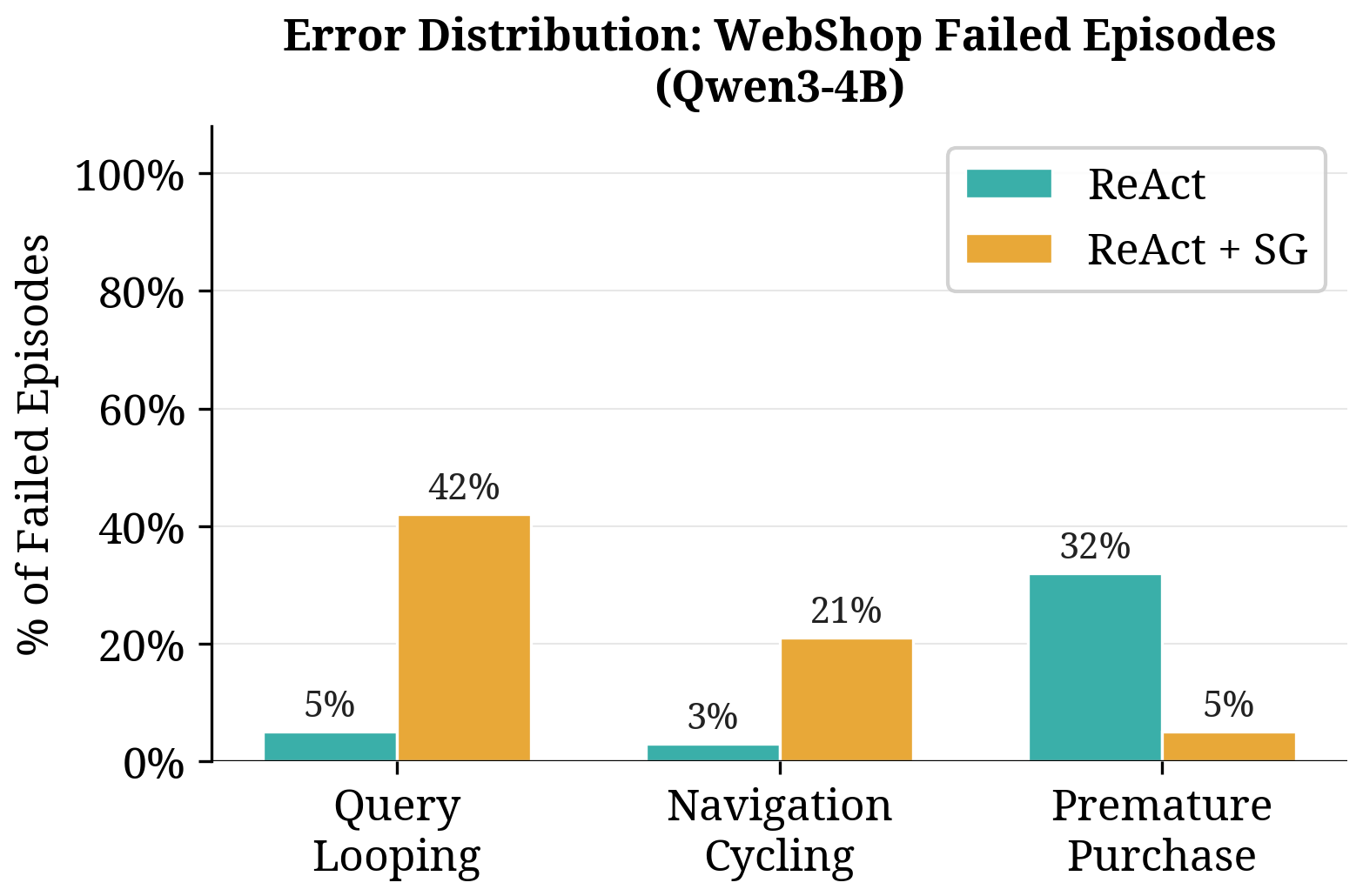}
        \label{fig:error_webshop}
    \end{subfigure}

    \vspace{0.5em}

    % Row 3: Model C
    \begin{subfigure}[b]{0.32\textwidth}
        \centering
        \includegraphics[width=\textwidth]{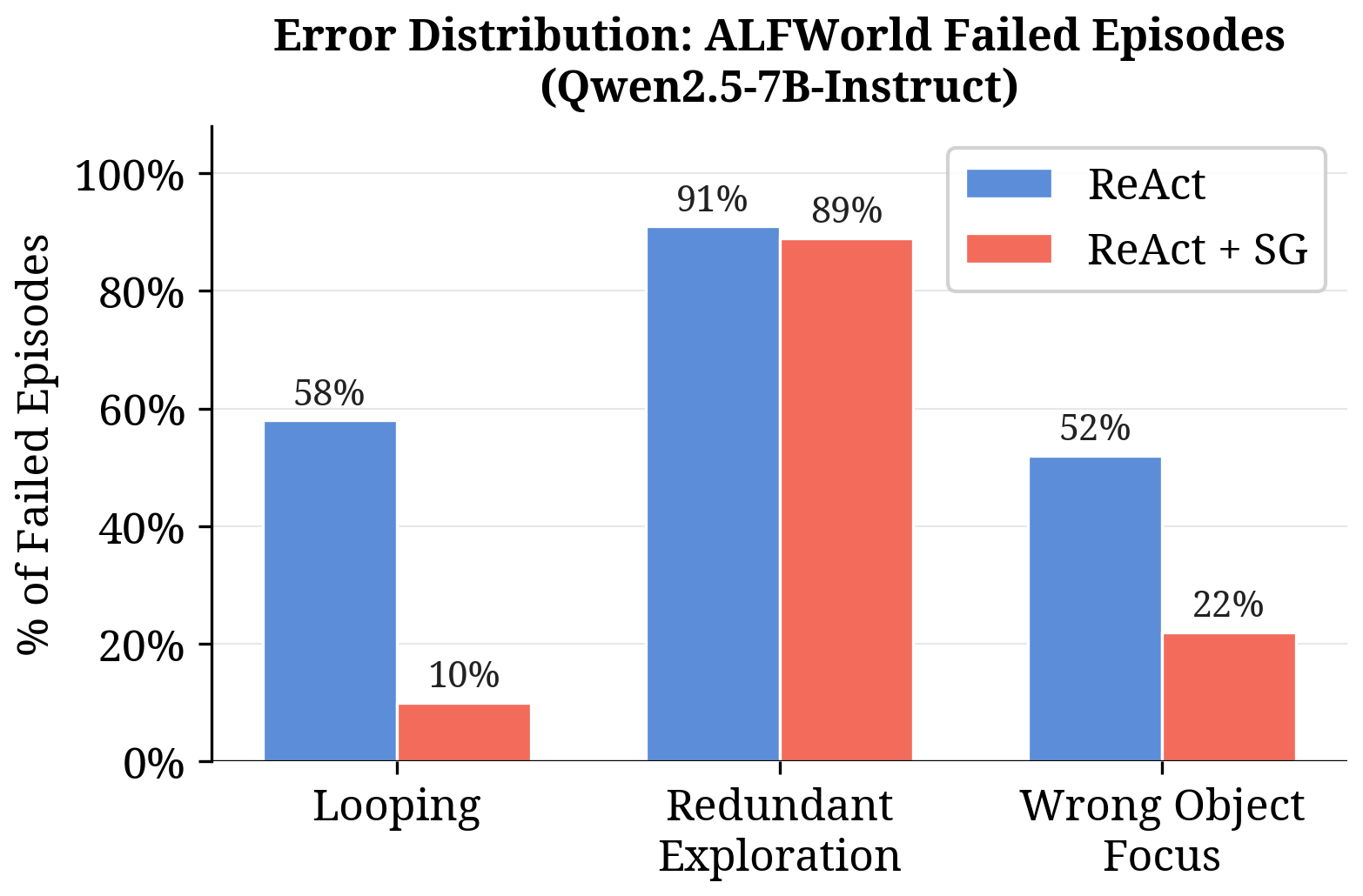}
    \end{subfigure}
    \hfill
    \begin{subfigure}[b]{0.32\textwidth}
        \centering
        \includegraphics[width=\textwidth]{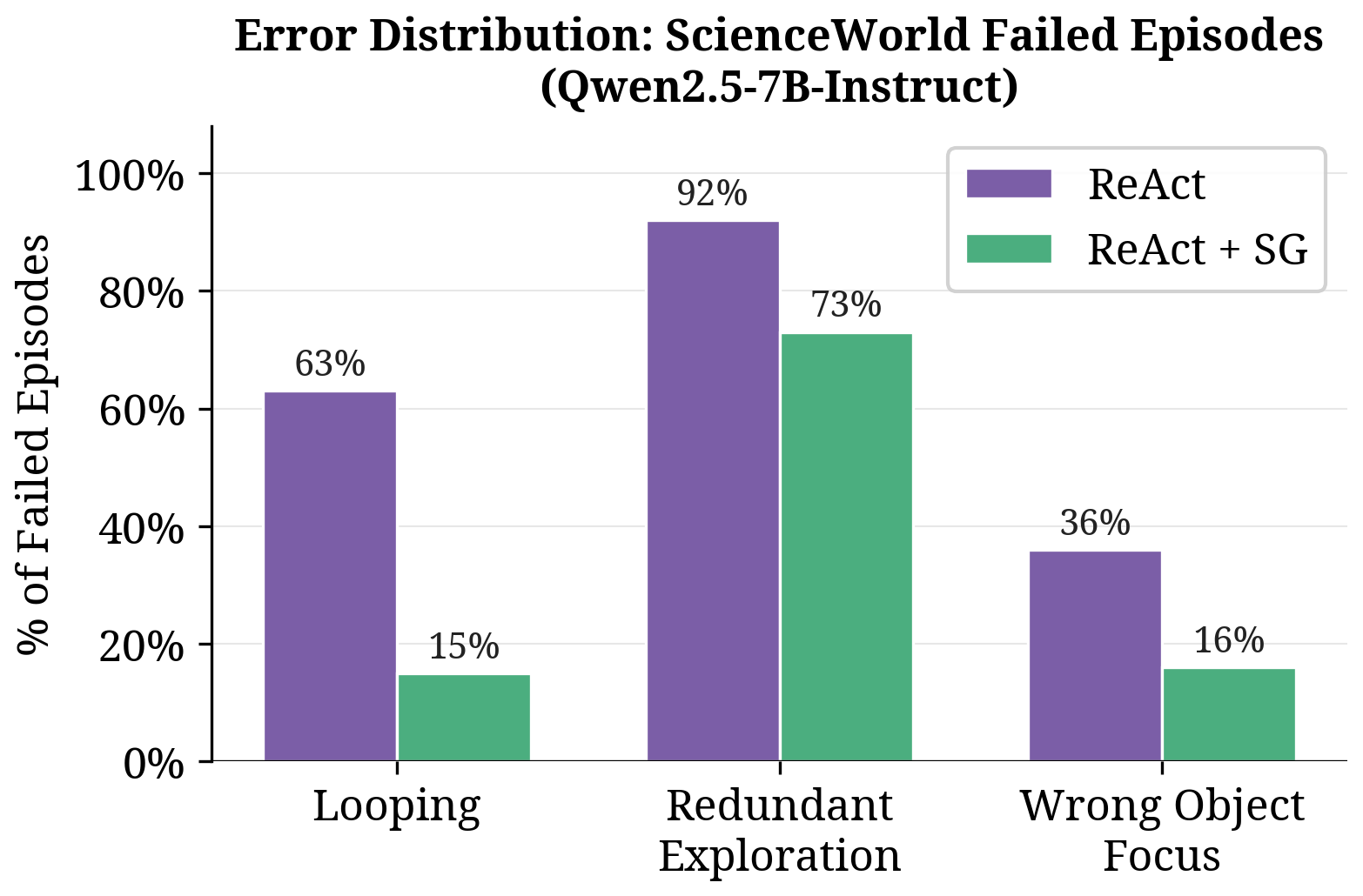}
    \end{subfigure}
    \hfill
    \begin{subfigure}[b]{0.32\textwidth}
        \centering
        \includegraphics[width=\textwidth]{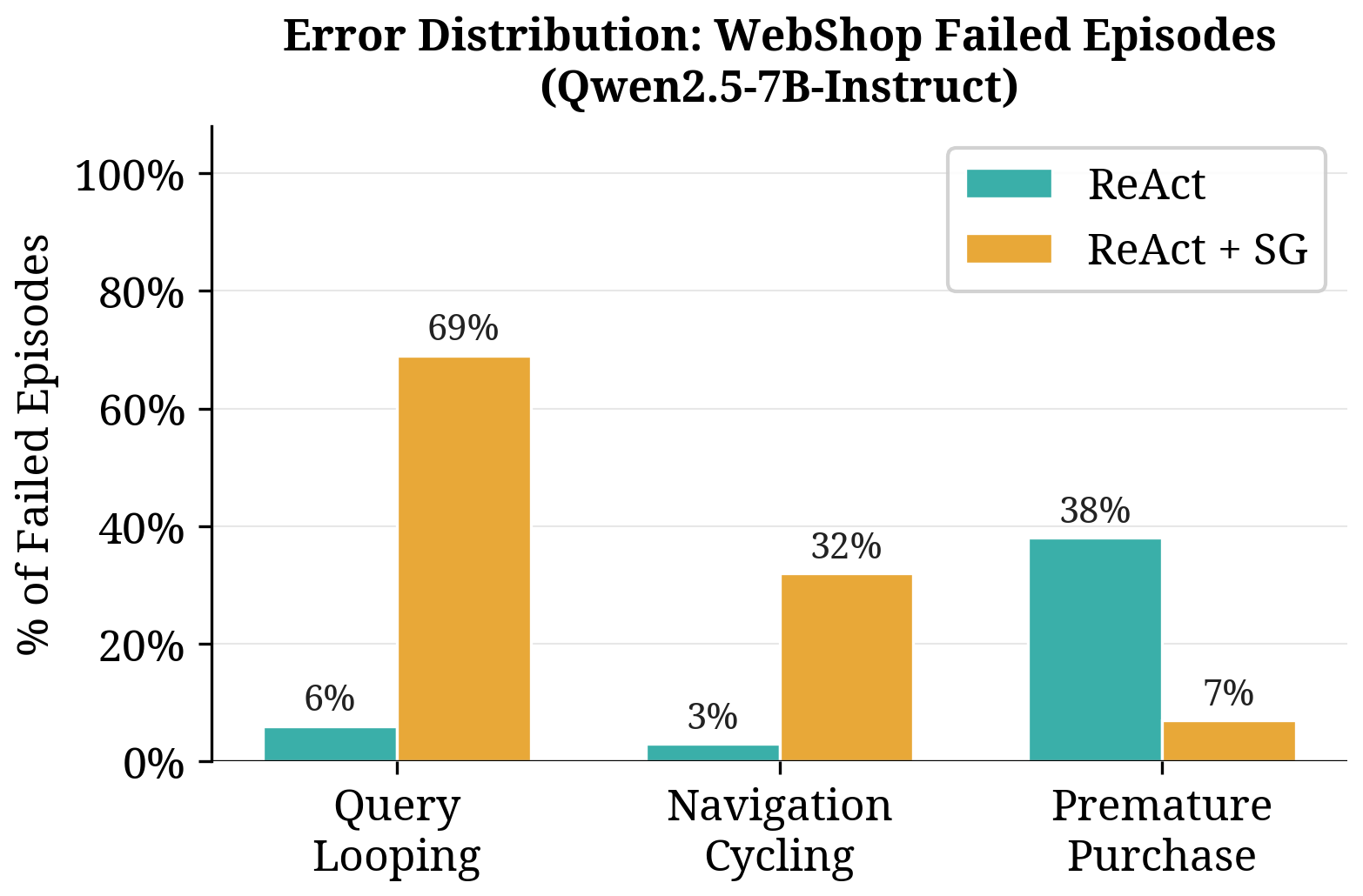}
    \end{subfigure}

    \caption{
        Error distribution among failed episodes across three benchmarks and three base models.
        Each bar shows the percentage of failed episodes exhibiting the corresponding error type;
        episodes may be counted in multiple categories.
    }
    \label{fig:error_analysis}
\end{figure}

\paragraph{Results.}
Figure~\ref{fig:error_analysis} reports error rates among failed episodes for each environment.
 
\textbf{ALFWorld.}
Self-guidance most substantially reduces looping (90.0\%\,$\to$\,38.5\%), with modest decreases in redundant exploration (51.5\%\,$\to$\,46.2\%) and wrong-object focus (52.3\%\,$\to$\,45.3\%).
This suggests self-guidance primarily acts as a \textit{loop-breaking signal}: it reliably detects and interrupts repetitive action cycles, while subtler errors such as misidentifying the target object remain largely unresolved.
 
\textbf{ScienceWorld.}
Self-guidance produces broader improvements across all three categories: looping (81.9\%\,$\to$\,17.5\%), redundant exploration (93.8\%\,$\to$\,57.1\%), and wrong-object focus (44.4\%\,$\to$\,24.6\%), accompanied by a success rate gain of 49.3\%\,$\to$\,58.0\%.
The more comprehensive error reduction relative to ALFWorld may reflect that ScienceWorld tasks require longer multi-step reasoning, giving the self-guidance model more opportunities to provide corrective feedback throughout the trajectory.
 
\textbf{WebShop.}
The pattern here is qualitatively different.
Presence of self-guidance sharply reduces premature purchases (69.1\%\,$\to$\,21.6\%), correctly penalizing episodes where the agent buys without verifying required attributes.
However, query looping increases substantially (17.5\%\,$\to$\,67.8\%). This is consistent with the behavior of a weak base model that is biased \emph{not} to act hastily but lacks the product-evaluation strategy needed to act correctly instead: faced with negative feedback on early purchase attempts, the agent falls back to repeatedly re-issuing search queries.

\section{Compatibility with Other RL Algorithm}
\label{sec:ablation_optimizer}
\begin{figure}[t]
    \centering
    \includegraphics[width=0.7\linewidth]{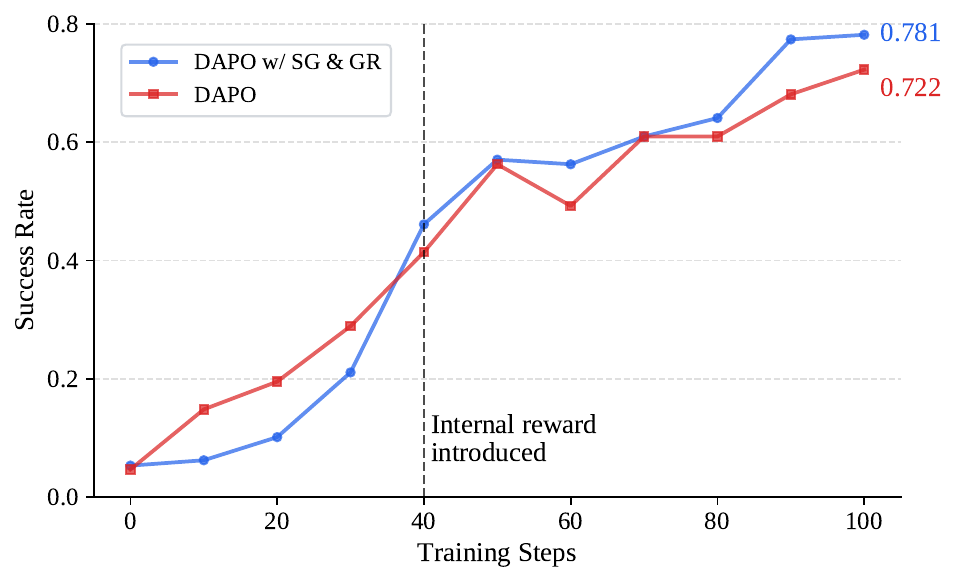}
    \caption{Training curves of DAPO and DAPO w/ \ourmethodfullshort{}. Our method achieves a final success rate of 78.1\% vs.\ 72.2\% for the DAPO baseline.}
    \label{fig:dapo_training_curve}
\end{figure}

Although we instantiate our method on top of GRPO throughout the main experiments, the proposed \ourmethodfull{} training mechanism is not specific to a single RL algorithm. Our method modifies the training signal by introducing self-guidance and an additional guidance reward, while leaving the underlying policy optimization algorithm unchanged. This makes it naturally compatible with other GRPO-family or outcome-driven RL algorithms.

To test this, we additionally apply our method on top of \textbf{DAPO}~\citep{yu2025dapo}, replacing vanilla GRPO with the DAPO objective while keeping the same \ourmethodfull{} design. We train Qwen3-1.7B on WebShop for \texttt{100} epochs with a group size of \texttt{8}, learning rate \texttt{\(1\times10^{-6}\)}, KL coefficient \texttt{0.01}, and DAPO clip ratios \texttt{(0.2,0.28)}; the guidance reward is introduced at step \texttt{40}. As shown in Figure~\ref{fig:dapo_training_curve}, once the guidance reward is introduced at step 40, our method begins to pull ahead of the DAPO baseline and maintains a higher success rate through the remainder of training, resulting in a stronger final performance. This suggests that the benefit of learned self-guidance and guidance reward is complementary to the choice of the underlying RL algorithm rather than being tied to GRPO alone.

\section{Details of Offline Self-Guidance Model Distillation}
\label{app:offline_rater_distillation}

This section provides additional details for the offline self-guidance distillation alternative discussed in Section~\ref{sec:offline_vs_online_rater}.

\subsection{Setup and Distillation Objective}

We conduct the offline distillation experiment on ALFWorld. The student self-guidance model is Qwen3-1.7B, and the teacher is a stronger Qwen3-32B model. For each trajectory prefix, the student observes only the prefix up to step $t$ and produces a rating for the current state. The teacher is given the full trajectory in hindsight and evaluates the student output using the complete rollout outcome. We instantiate both teacher and student within the Qwen3 family to reduce additional mismatch from heterogeneous model families, so that the main gap comes from context and capability rather than architecture differences.

\begin{figure*}[t]
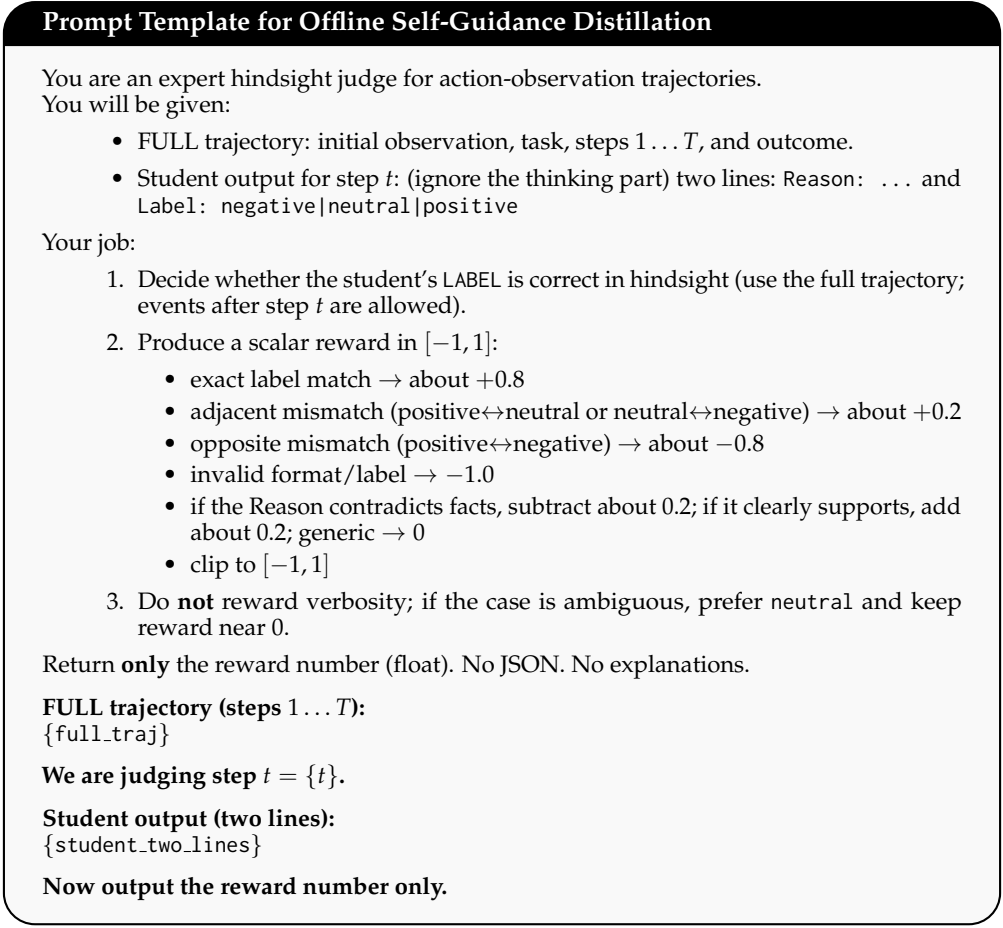

\centering
\begin{tcolorbox}[
    colback=gray!5,
    colframe=black,
    boxrule=0.8pt,
    arc=4mm,
    width=0.95\textwidth,
    title={Prompt Template for Offline Self-Guidance Distillation},
    fonttitle=\bfseries
]
    \small
    You are an expert hindsight judge for action-observation trajectories.
    
    You will be given:
    \begin{itemize}
        \item FULL trajectory: initial observation, task, steps $1 \dots T$, and outcome.
        \item Student output for step $t$: (ignore the thinking part) two lines: \texttt{Reason: ...} and \texttt{Label: negative|neutral|positive}
    \end{itemize}
    
    Your job:
    \begin{enumerate}
        \item Decide whether the student's \texttt{LABEL} is correct in hindsight (use the full trajectory; events after step $t$ are allowed).
        \item Produce a scalar reward in $[-1,1]$:
        \begin{itemize}
            \item exact label match $\rightarrow$ about $+0.8$
            \item adjacent mismatch (positive$\leftrightarrow$neutral or neutral$\leftrightarrow$negative) $\rightarrow$ about $+0.2$
            \item opposite mismatch (positive$\leftrightarrow$negative) $\rightarrow$ about $-0.8$
            \item invalid format/label $\rightarrow -1.0$
            \item if the Reason contradicts facts, subtract about $0.2$; if it clearly supports, add about $0.2$; generic $\rightarrow 0$
            \item clip to $[-1,1]$
        \end{itemize}
        \item Do \textbf{not} reward verbosity; if the case is ambiguous, prefer \texttt{neutral} and keep reward near $0$.
    \end{enumerate}
    
    Return \textbf{only} the reward number (float). No JSON. No explanations.
    
    \medskip
    \noindent\textbf{FULL trajectory (steps $1 \dots T$):} \\
    \texttt{\{full\_traj\}}
    
    \medskip
    \noindent\textbf{We are judging step $t=\{t\}$.}
    
    \medskip
    \noindent\textbf{Student output (two lines):} \\
    \texttt{\{student\_two\_lines\}}
    
    \medskip
    \noindent\textbf{Now output the reward number only.}
\end{tcolorbox}
\caption{Prompt template used for offline self-guidance model distillation.}
\label{fig:offline_rater_prompt}
\end{figure*}

For each trajectory prefix, the student produces a rating using the same prompt defined in~\ref{app:experiment_details}. The teacher scores the student output using the full trajectory in hindsight and returns a scalar reward in $[-1,1]$, which serves as the supervision signal for offline distillation. The exact prompt template is shown in Figure~\ref{fig:offline_rater_prompt}.

\subsection{Co-Evolving Improves the Self-Guidance Model}
\label{app:coevolving_rater}
We further examine whether co-evolving the self-guidance model with the policy improves the self-guidance model itself, rather than only the policy. To isolate this effect, we experiment on Webshop using Qwen3-1.7B. We fix the policy to \textbf{checkpoint 40} and vary only the self-guidance model across different training checkpoints. If co-evolving indeed strengthens the self-guidance model, then later checkpoints should provide better guidance even for the same fixed policy.

Table~\ref{tab:coevolving_rater_ckpt} confirms this trend. Using \textbf{checkpoint 10} as the self-guidance model yields a score of \textbf{39.8} and a success rate of \textbf{18.0\%}, while replacing it with \textbf{checkpoint 40} improves performance to \textbf{51.2} and \textbf{25.0\%}. The best result is obtained with \textbf{checkpoint 80}, which reaches a score of \textbf{60.5} and a success rate of \textbf{27.0\%}. Since the policy is held fixed throughout this comparison, these gains cannot be attributed to policy improvement. Instead, they indicate that co-evolving during training progressively improves the self-guidance model itself.

\begin{table}[t]
    \centering
    \caption{Effect of self-guidance model checkpoints on a fixed policy. The policy is fixed to \textbf{ckpt40}, while the self-guidance model is varied across training checkpoints.}
    \label{tab:coevolving_rater_ckpt}
    \begin{tabular}{lcc}
        \toprule
        Self-Guidance Model & Score & Success Rate (\%) \\
        \midrule
        checkpoint 10 & 39.8 & 18.0 \\
        checkpoint 40 & 51.2 & 25.0 \\
        checkpoint 80 & \textbf{60.5} & \textbf{27.0} \\
        \bottomrule
    \end{tabular}
\end{table}

\end{document}